\documentclass{article}

\newcommand{\status}{0}
\newcommand{\Description}[1]{#1}

\usepackage[accepted]{tmlr}

\usepackage[utf8]{inputenc} 
\usepackage[T1]{fontenc}    
\usepackage{hyperref}       
\hypersetup{
    colorlinks=true,
    citecolor=blue!50!black,
    linkcolor=brown!50!black,
    filecolor=magenta,      
    urlcolor=cyan,
    }
\usepackage{url}            
\usepackage{booktabs}       
\usepackage{amsfonts}       
\usepackage{nicefrac}       
\usepackage{microtype}      
\usepackage{mathtools}
\usepackage[table]{xcolor}
\usepackage{tikz}
\usetikzlibrary{fit,shapes.geometric}
\usepackage{floatrow}
\usepackage{float}

\input{packages}
\definecolor{INCOMPLETECOLOR}{RGB}{178,34,34}
\definecolor{UNDERREVISIONCOLOR}{RGB}{210,121,121}
\definecolor{FEEDBACKNEEDEDCOLOR}{RGB}{230,170,50}
\definecolor{FEEDBACKGIVENCOLOR}{RGB}{121,210,121}
\definecolor{COMPLETECOLOR}{RGB}{121,124,210}
\definecolor{QINCHANCOLOR}{RGB}{0,190,255}

\definecolor{TODOCOLOR}{RGB}{255,0,0}
\definecolor{MONDECOLOR}{RGB}{0,0,255}
\definecolor{QICOLOR}{RGB}{200,10,0}
\definecolor{ANJULCOLOR}{RGB}{127,127,0}
\definecolor{RACHELCOLOR}{RGB}{127,0,127}
\definecolor{KENNYCOLOR}{RGB}{255,127,127}
\definecolor{GUESTCOLOR}{RGB}{0,127,127}

\definecolor{WHITE}{RGB}{255,255,255}

\newcommand{\nothing}[1]{}
\newcommand{\isolated}[1]{\hfill\break#1\xspace}
\newcommand{\Caption}[2]{\caption[#1]{{\em #1} #2}}

\newcommand{\kenny}[1]{[\hl{\textbf{K}}: \textcolor{orange!60!black}{#1}]}

\newcommand{\checkagain}[1]{}
\newcommand{\wrm}[1]{}
\newcommand{\wblank}[1]{}

\ifthenelse{\equal{\status}{0}} {                     

    \newcommand{\todo}[1]{%
        \addcontentsline{toc}{subsection}{
            \protect\numberline{}
            \textcolor{TODOCOLOR}{[TODO] #1}}
            \textcolor{TODOCOLOR}{[TODO] \emph{#1}}}
    \newcommand{\warning}[1]{\todo{#1}}
    \newcommand{\note}[1]{{\it\color{blue} #1}}

    \newcommandx{\monde}[2][1=]
        {\setulcolor{MONDECOLOR}{\ul{#1}}
         \isolated{\textcolor{MONDECOLOR}{\textbf{Monde:} #2}}}
    \newcommandx{\qisun}[2][1=]
        {\setulcolor{QICOLOR}{\ul{#1}}
         \isolated{\textcolor{QICOLOR}{[\textbf{Q:} #2]}}}
    \newcommandx{\anjul}[2][1=]
        {\setulcolor{ANJULCOLOR}{\ul{#1}}
         \isolated{\textcolor{ANJULCOLOR}{\textbf{Anjul:} #2}}}
    \newcommandx{\rachel}[2][1=]
        {\setulcolor{RACHELCOLOR}{\ul{#1}}
         \isolated{\textcolor{RACHELCOLOR}{\textbf{Rachel:} #2}}}
    \newcommandx{\qinchan}[2][1=]
        {\setulcolor{QINCHANCOLOR}{\ul{#1}}
         \isolated{\textcolor{QINCHANCOLOR}{\textbf{Qinchan (Wing):} #2}}}

    \newcommandx{\guest}[3][1=]
        {\setulcolor{GUESTCOLOR}{\ul{#1}} \textcolor{GUESTCOLOR}
        {[\textbf{#2:} #3]}}
}{                                                    
    \newcommand{\todo}[1]{}
    \newcommand{\warning}[1]{}
    \newcommand{\note}[1]{}

    \newcommandx{\monde}[2][1=]{#1}
    \newcommandx{\qinchan}[2][1=]{#1}
    \newcommandx{\qisun}[2][1=]{#1}
    \newcommandx{\kenny}[2][1=]{#1}
    \newcommandx{\guest}[3][1=]{#1}
}

\ifthenelse{\equal{\status}{0}} {          
    
}{}
\ifthenelse{\equal{\status}{1}} {          
    
}{}
\ifthenelse{\equal{\status}{2}} {          
    
}{}
\ifthenelse{\equal{\status}{3}} {          
    
}{}
\ifthenelse{\equal{\status}{4}} {          
    
}{}

\ifthenelse{\equal{\status}{0}} {
    \newcommand{\badge}[2]{\colorbox{#1}{\small\textcolor{WHITE}{\texttt{#2}}}}
    \newcommand{\headerBadge}[2]{\hspace*{\fill}\badge{#1}{#2}}

}{
    \newcommand{\badge}[2]{}{}
    \newcommand{\headerBadge}[2]{}{}

}

\def\bx{\mathbf{x}}%
\def\by{\mathbf{y}}%
\def\bi{\mathbf{i}}%
\def\defeq{\stackrel{.}{=}}%
\def\1{\boldsymbol{1}}%

\newcommand{\methodName}{\textsl{CATImage}\xspace}

\newcommand{\ourtitle}{Cost-Aware Routing for Efficient Text-To-Image Generation}
\title{\ourtitle}
\usepackage{environ}

\NewEnviron{revision}{%
    \color{blue}%
    \BODY
    \color{black}%
}

\newcommand{\ssp}{3pt}
\author{\name Qinchan (Wing) Li$^\diamond$ \email ql840@nyu.edu \\[\ssp]
      \name Kenneth Chen$^\diamond$ \email kc4906@nyu.edu \\[\ssp]
      \name Changyue (Tina) Su$^\diamond$ \email cs7483@nyu.edu\\[\ssp]
      \name Wittawat Jitkrittum$^\dagger$\thanks{Now at Eigen 4D Inc. (\url{https://eigen4d.com}).} \email wittawatj@gmail.com\\[\ssp]
      \name Qi Sun$^\diamond$ \email qisun@nyu.edu\\[\ssp]
      \name Patsorn Sangkloy$^{\diamond*}$ \email patsorn.sangkloy@gmail.com \AND
      \addr $\diamond$ Tandon School of Engineering, New York University \\
      \addr $\dagger$ Google, New York  
      }

\def\month{03}  
\def\year{2026} 
\def\openreview{\url{https://openreview.net/forum?id=Jbe9AVsYS6}} 

\begin{document}

\maketitle

\begin{abstract}
Diffusion models are well known for their ability to  generate a high-fidelity image for an input prompt through an iterative denoising process.
Unfortunately, the high fidelity also comes at a high computational cost due to the inherently sequential generative process.
In this work, we seek to optimally balance quality and computational cost, and propose a framework to allow the amount of computation to vary for each prompt, depending on its complexity. 
Each prompt is automatically routed to the most appropriate text-to-image generation function, which may correspond to a distinct number of denoising steps of a diffusion model, or a disparate, independent text-to-image model. 
Unlike uniform cost reduction techniques (e.g., distillation, model quantization), our approach achieves the optimal trade-off by learning  to reserve expensive choices (e.g., 100+ denoising steps) only for a few complex prompts, and employ more economical choices (e.g., small distilled model) for less sophisticated prompts. 
We empirically demonstrate on COCO and DiffusionDB that by learning to route to nine already-trained text-to-image models, our approach is able to deliver an average quality that is higher than that achievable by any of these models alone. 
Code is available at \url{https://github.com/winglicopy/CATImage}.
\end{abstract}



\section{Introduction}
\label{sec:introduction}


While diffusion models have set a new standard for photorealism in generative art \citep{ho2020denoising}, their operational costs remain a major challenge. The generation of a single image can involve many denoising steps, each utilizes a learned denoiser model with potentially over a billion parameters \citep{rombach2022high}. This makes in-the-wild adoption (i.e., on-device) challenging and raises valid concerns about their environmental sustainability \citep{genaiEnergy,genaiEnergyNature,kaack2022aligning}. To address this, a significant body of research has explored optimization strategies such as network simplification \citep{li2024snapfusion,li2023autodiffusion} and model distillation \citep{sauer2024fast, salimans2022progressive, meng2023distillation, liu2023instaflow}.

\begin{figure}[t]
\centering
\subfloat[\textit{`a white and empty wall'}]{
    \includegraphics[width=0.28\linewidth]{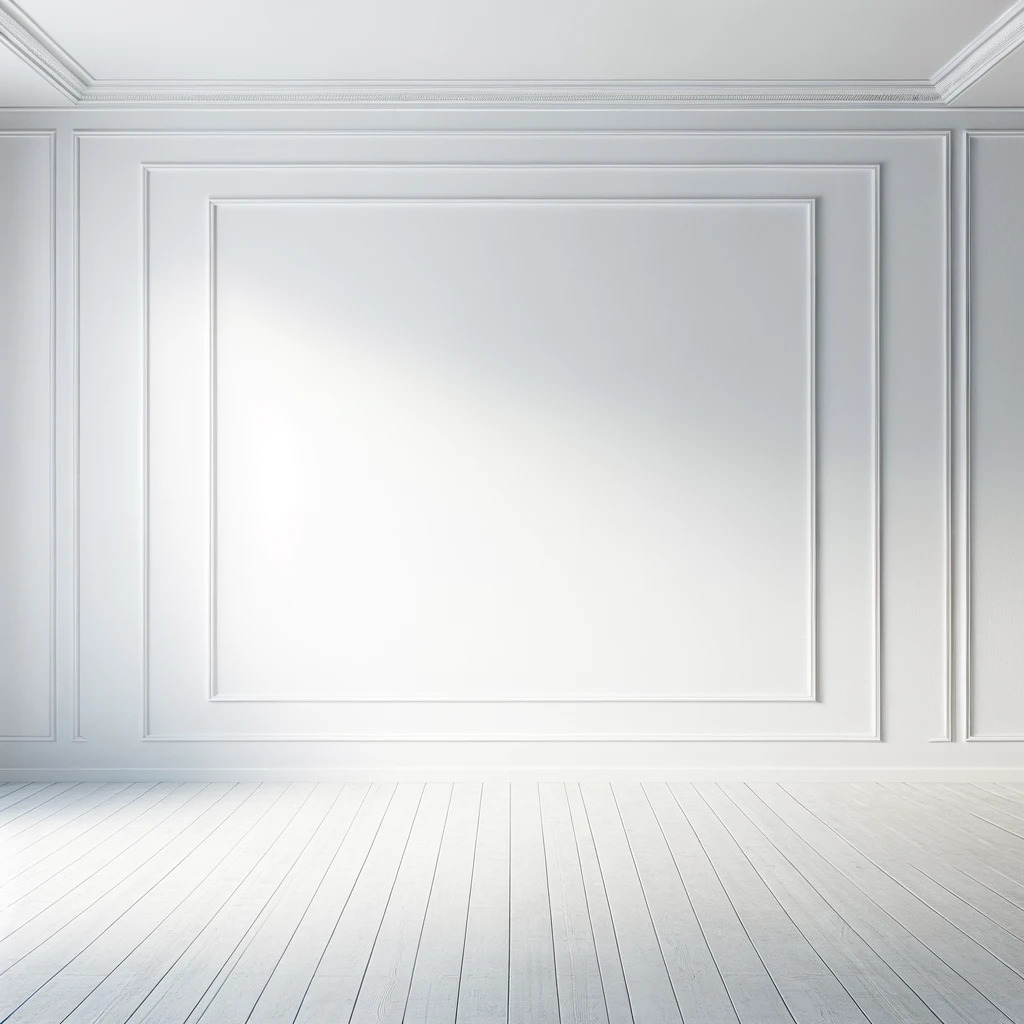}
    \label{fig:intro:simple}
}
\subfloat[\textit{`a colorful park with a crowd'}]{
    \includegraphics[width=0.28\linewidth]{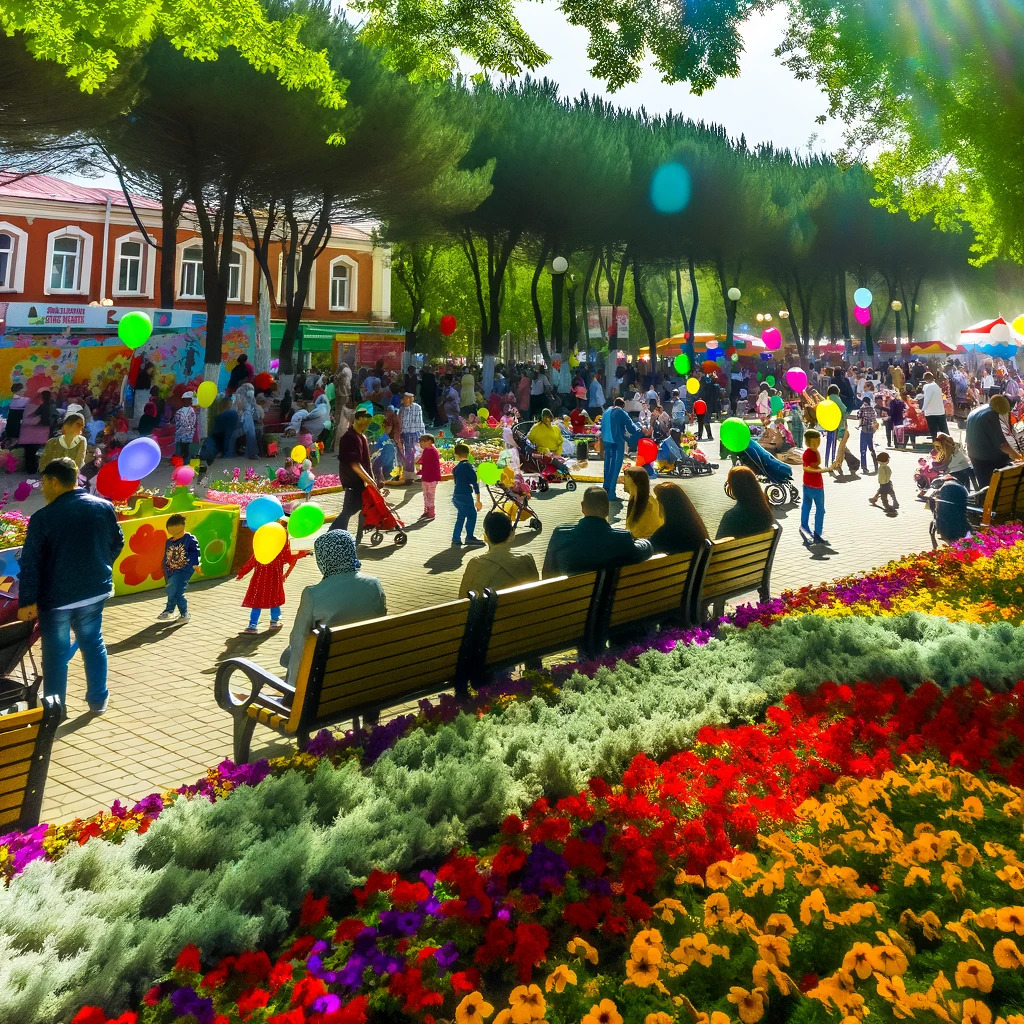}
     \label{fig:intro:complex}
}
\hspace{0.02\linewidth}%
\subfloat[\textit{quality trends across numbers of steps}]{
    \includegraphics[width=0.35\linewidth]{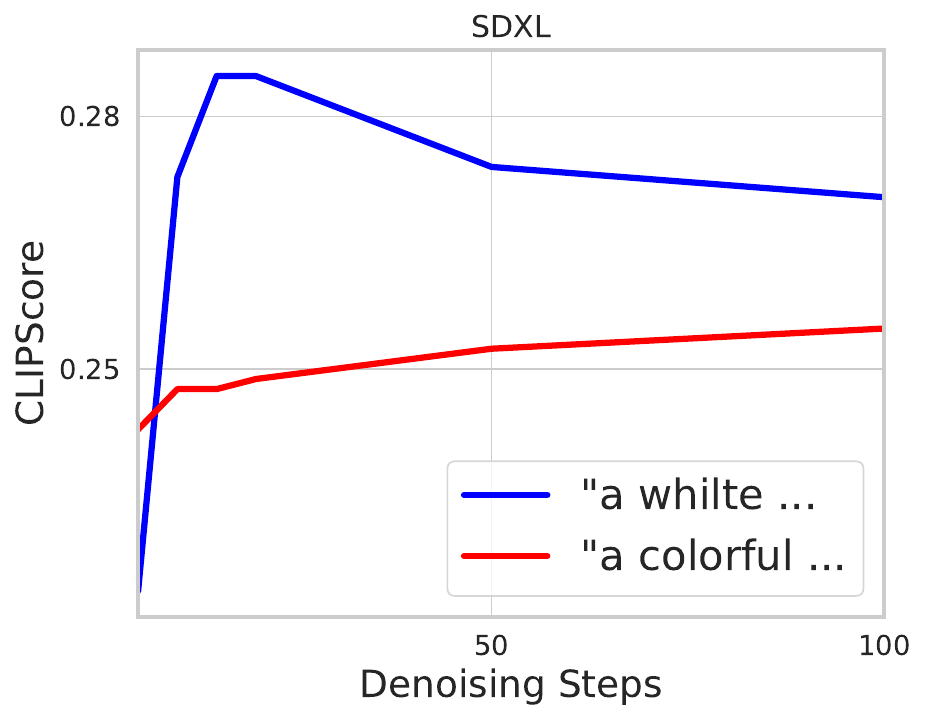}
     \label{fig:intro:curve}
}
\Caption{
    Two input prompts that require different denoising steps to ensure quality. }
{%
As shown in \Cref{fig:intro:curve}, prompt \Cref{fig:intro:simple} only requires a small number of denoising steps to reach a high CLIPScore. By contrast, the more complex prompt \Cref{fig:intro:complex} requires over 100 steps to reach a similar quality. 
Key to our proposed \methodName is to allocate an appropriate amount of computation for each prompt, so that the overall computational cost is reduced while the quality remains the same.
} 
\label{fig:intro}
\end{figure}

However, these existing methods typically apply the same degree of optimization irrespective of the task's intrinsic difficulty. This results in a single model with a fixed computational cost, which is inherently suboptimal as the generative effort required to synthesize an image varies with the complexity of the input prompt.
For example, a simple prompt like \textit{a white and empty wall} requires fewer denoising steps to generate a high-quality image than a complex one like \textit{a colorful park with a crowd}, as shown in \Cref{fig:intro}.

With the motivation to adaptively allocate computational budget,  we present \methodName, a framework that allows the amount of computation for text-to-image generation to vary for each prompt.
Our framework operates with a pre-defined set of choices that can be chosen adaptively for each input prompt. Each choice represents a text-to-image generation function and has a distinct profile of computational cost and the expected image quality. 
Concretely, these choices may correspond to different numbers of denoising steps of the same diffusion model (i.e., homogeneous choices),   disparate, independent text-to-image generative models (i.e., heterogeneous choices), or a combination of both.
The proposed \methodName aims to adaptively select the right choice (i.e., ``routing'') for each input prompt, in such a way that expensive choices (e.g., 100+ denoising steps) are  reserved only for complex prompts. 
Our approach enables a joint deployment of diverse text-to-image models and has a potential to deliver higher average image quality compared to using any individual model in the pool, while allowing the average computational cost to be adapted at deployment time.

%
%
%
In summary, our contributions are as follows.
\begin{enumerate}[leftmargin=*]
    \item We precisely formulate a constrained optimization problem for the above routing problem (\Cref{sec:problem}). The formulation aims to maximize average image quality subject to a budget constraint on the generation cost.
    

    \item We study the theoretically optimal routing rule that optimally trades off the average quality and cost (\Cref{sec:opt_rule}). Based on the optimal rule, we construct a plug-in estimator that can be trained from data.

    \item We perform a series of objective analyses on the COCO \citep{lin2014microsoft} and DiffusionDB datasets \citep{wangDiffusionDBLargescalePrompt2022}. Our findings show that, through adaptive routing, our proposal matches the quality of the largest model in the serving pool (namely, Stable Diffusion XL from \citet{radford2021learning} with 100 denoising steps) with only a fraction of its computational cost (\Cref{tab:cost_ratio}).\footnote{We will release the code and data upon paper publication.}
    
\end{enumerate}

\section{Background: Text-To-Image Generative Models}
\label{sec:background} \nothing{In this section, we give a brief introduction
to text-to-image generative models in general, focusing on diffusion
models as a specific instance. This section serves as a precursor to our proposal in \Cref{sec:main}.}

\label{sec:text_to_image} Let $\bx\in\mathscr{X}$ denote an input
text prompt, and $\bi\in\mathscr{I}\defeq[0,1]^{W\times H\times3}$
denote an image described by the prompt, where $W,H\in\mathbb{N}$
denote the width and the height of the image (in pixels), and the
last dimension denotes the number of color channels. A text-to-image
generative model is a stochastic map $h\colon\mathscr{X}\to\mathscr{I}$
that takes a prompt $\bx$ as input and generates an image $h(\bx)\in\mathscr{I}$
that fits the description in the prompt $\bx$.
There are many model classes one may use to construct such a model
$h$, including conditional Generative Adversarial Networks (GANs)
\citep{ZhaXuLi2017,GooPouMir2014}, Variational Auto-Encoder (VAE)
\citep{KinWel2022}, and diffusion models \citep{ho2020denoising},
among others. 

\textbf{Diffusion models}
A specific class of text-to-image generative models that has recently
been shown to produce high-fidelity images is given by diffusion-based
models \citep{SahChaSax2022,ho2020denoising,HoSahCha2022}. A diffusion
generative model relies on a function $g\colon\mathscr{X}\times\mathbb{N}\times\mathbb{R}^{D}\to\mathscr{I}$
that takes as input a prompt $\bx,$ the number of denoising steps
$T\in\mathbb{N}$, a noise vector $\mathbf{z}\in\mathbb{R}^{D}$ with
$D=3\cdot WH$, and generates an image $\bi=g(\bx,T,\mathbf{z})$.
Image generation is done by iteratively refining the initial noise
vector $\mathbf{z}$ for $T$ iterations to produce the final image.
The noise vector $\mathbf{z}\sim\mathcal{N}(\mathbf{0},\mathbf{I})$
is typically sampled from the standard multivariate normal distribution
and the $T$ refinement steps correspond to the reverse diffusion
process, which reconstructs an image from a random initial state \citep{ho2020denoising}.
With $\mathbf{z}\sim\mathcal{N}(\mathbf{0},\mathbf{I})$ understood
to be an implicit source of randomness, we define $h_{T}(\bx)\defeq g(\bx,T,\mathbf{z})$
to be an image sampled from the diffusion model using $T$ diffusion
steps. With $T$ chosen, $h_{T}\colon\mathscr{X}\to\mathscr{I}$ is
thus an instance of text-to-image generative models as described earlier. The importance of this view will be apparent
when we describe our proposed method in \Cref{sec:main}, which enables
an automatic selection of the number of denoising steps, separately for each
prompt. Typically, the number of denoising steps is pre-chosen according to the computational budget available at inference time, with a low
value of $T$, giving a lower computational cost at the expense of
image quality.

\begin{figure}
  \floatbox[{\capbeside\thisfloatsetup{capbesideposition={right,center},capbesidewidth=5.2cm}}]{figure}[\FBwidth]
  {\Caption{ 
Our pipeline.}{
During training (dashed box), a quality estimator is trained to predict per-prompt quality scores for all routing candidates $h^{(1)}, \ldots, h^{(M)}$. 
At inference time (bottom), given a prompt, predicted quality scores of all routing candidates are adjusted by their respective costs. 
The routing candidate that has the highest cost-adjusted score is chosen (see Eq. \eqref{eq:emp_rule}).
}\label{fig:pipeline}}
  {\includegraphics[width=0.5\textwidth]{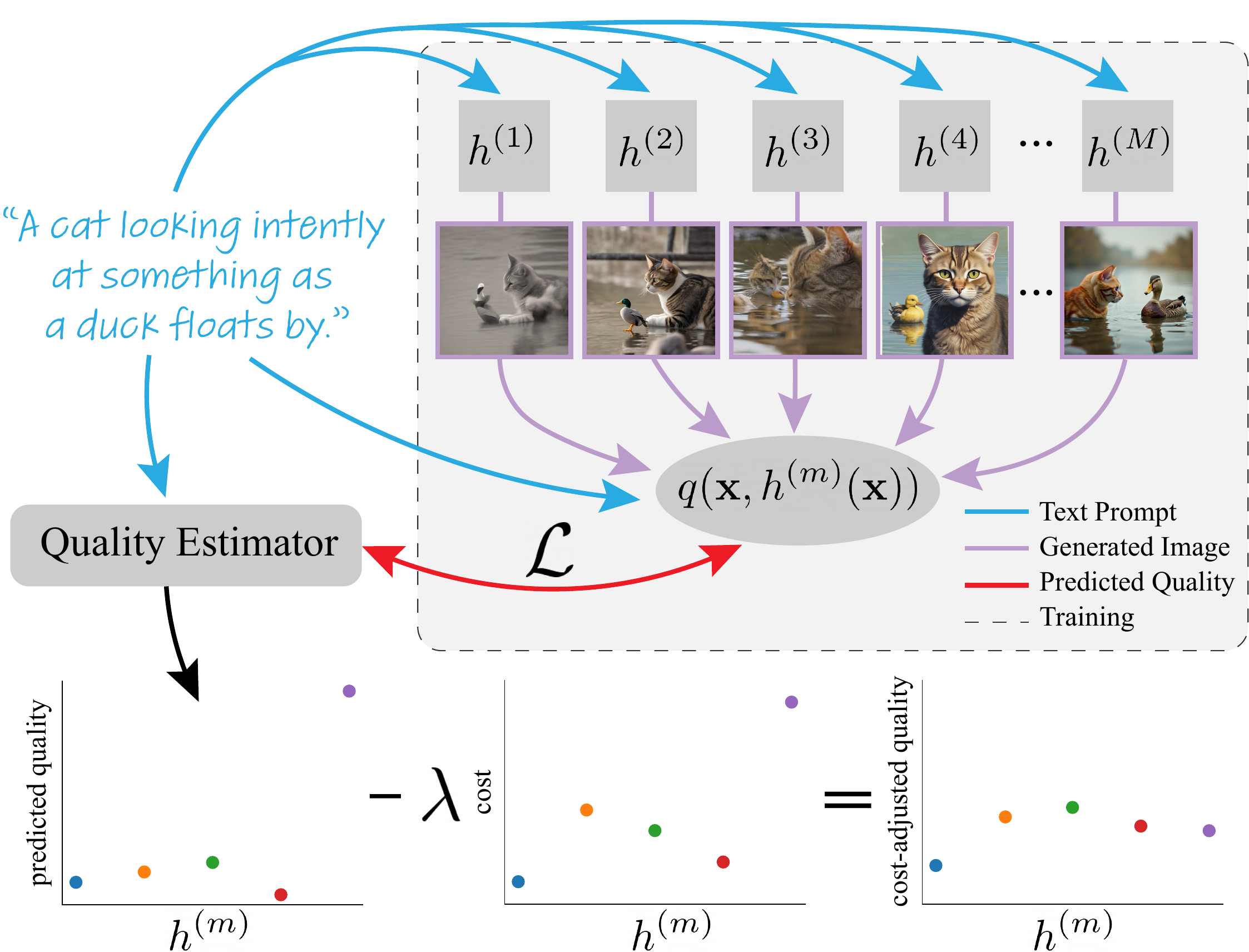}}
  \vspace{-4mm}
\end{figure}

\section{Cost-Aware Text-To-Image Generation}
\label{sec:main} We now describe our main proposal termed \methodName{}
(\uline{C}ost-\uline{A}ware \uline{T}ext-based \uline{Ima}ge \uline{Ge}neration),
which seeks to minimize inference cost by adaptively adjusting the
cost per prompt, depending on its complexity. As illustrated
in \Cref{fig:intro}, in the case of a diffusion model, our
key observation is that not all prompts require a large number of
denoising steps to ensure quality. Thus, inference efficiency can
be achieved by spending a small amount of computation for easy prompts.
Our proposed framework is general and allows cost adjustment in a
per-prompt manner via selecting an appropriate amount of resources
from homogeneous choices (i.e., adaptively varying the number of denoising
steps of a single diffusion model), or heterogeneous choices (i.e.,
adaptively route prompts to disparate, independent generative models). 

We start by formalizing the cost-aware text-to-image generation task
as a learning-to-route problem in \Cref{sec:problem}. The formulation
can be theoretically shown (\Cref{sec:opt_rule}) to have a simple
Bayes optimal routing rule, involving subtracting off the expected
quality metrics with the costs of candidate numbers of denoising steps. We show
 that the optimal rule can be estimated from
data, and propose two estimators: a Transformer-based estimator \citep{VasShaPar2017},
and a $K$-nearest neighbors (KNN) model.

\subsection{Problem Formulation}

\label{sec:problem}Let $[n]\defeq\{1,2,\ldots,n\}$ denote the set
of counting numbers up to $n$. Suppose that we are given a fixed
set of $M$ choices $\mathscr{H}\defeq\{h^{(1)},\ldots,h^{(M)}\}$
where each choice $h^{(i)}\colon\mathscr{X}\to\mathscr{I}$ represents
a trained generative model (see \Cref{sec:text_to_image} for a precise
definition). Our goal is to derive a routing rule that optimally (in
the sense of quality-cost trade-offs) chooses the best model to invoke
for each input prompt. These $M$ \emph{base models} may be homogeneous,
being derived from a single diffusion model with varying numbers of diffusion
steps, a mix of heterogeneous generative model classes, or a combination
of both. For example, if we want to decide whether to use 20 or 50
number of denoising steps in the Stable Diffusion XL (SDXL)
model \citep{PodEngLac2023}, then $M=2$, and $\mathscr{H}=\{h^{(1)},h^{(2)}\}$
where the two models are both SDXL with the number of denoising steps fixed to 20
and 50, respectively. We will abstract away the details of the underlying
$M$ base models and propose a general framework that supports both
the homogeneous and heterogeneous cases (as shown in our experiments
in \Cref{sec:experiments}).

Suppose we are given a quality metric of interest $q\colon\mathscr{X}\times\mathscr{I}\to\mathbb{R}$
(see \textit{Quality Metrics} under \Cref{sec:exp:setting}), which takes as input a prompt-image
tuple, and estimates a quality score. We seek a router $r\colon\mathscr{X}\to[M]$
that predicts the index of the $M$ choices from a given prompt. We posit two desirable properties that the router ought to possess:
\begin{enumerate}[leftmargin=*]
\item The router must respect a specified budget constraint on the inference
cost.
\item Routing prompts to candidates in $\mathscr{H}$ must maximize average quality metric.
\end{enumerate}

Following similar formulations considered in \citet{JitGupMen2023,JitNarRaw2025,MaoMohMoh2023,TaiPatVer2024},
the above desiderata may be realized as a constrained optimization
problem:
%
\begin{align}
& \max_{r}Q(r)  \text{\thinspace\thinspace\ subject to\thinspace\thinspace\thinspace}C(r) \le B, 
 \quad \text{where}\label{eq:optimization} \\
Q(r) & \defeq\mathbb{E}\left[\sum_{m\in[M]}\1\left[r(\bx)=m\right]\cdot q(\mathbf{x},h^{(m)}(\bx))\right], \text{ and } \, \,
C(r) \defeq\mathbb{E}\left[\sum_{m\in[M]}\1\left[r(\bx)=m\right]\cdot c^{(m)}\right], \label{eq:avg_qc} 
\end{align}
where for $m\in[M]$, $c^{(m)}\ge0$ denotes the cost for the model
$h^{(m)}$ to produce one image for a given prompt, $\mathbb{E}$
denotes the expectation with respect to the population joint distribution
on all random variables (i.e., prompt $\bx$, and the sampled output
of $h^{(m)}$), $B\ge0$ is a hyperparameter specifying an upper bound
on the average cost. The optimization problem (\ref{eq:optimization})
thus seeks a router $r$ that maximizes the average quality $Q(r)$
subject to the constraint that the average cost (over all prompts)
is bounded above by $B$.
\begin{rem*}
The optimization problem is general and allows the per-model costs
to be in any unit suitable for the application (e.g., latency in seconds,
FLOP counts). Further, no practical constraint is imposed on the quality
metric function $q$. For instance, $q$ could be the CLIP score \citep{radford2021learning}.
Intuitively, if the budget $B$ is large, the cost constraint $C(r)\le B$
would have little effect, and the optimal router is expected to route
each prompt to the base model that can produce the highest quality metric score,
disregarding the cost of the model. In practice, such a model is often
the largest one in the pool $\mathscr{H}$, or the diffusion model
with the largest number of denoising steps. On the contrary, if $B$ is small,
the router would prioritize cost over quality, preferring to choose
a small base model (or a small number of denoising steps) over a larger
candidate. This proposal offers a framework to allow trading off average
quality with cost in a unified way by varying $B$. 
\end{rem*}



\subsection{Theoretically Optimal Routing Rule}
\label{sec:opt_rule}
Having formulated the constrained problem in
(\ref{eq:optimization}), we now investigate its theoretically optimal
solution. We will use the optimal solution to guide us on how to design
a practical router. Based on the results in \citet{JitGupMen2023,JitNarRaw2025},
the optimal solution to (\ref{eq:optimization}) is shown in Proposition\,\ref{prop:opt_router}.
\begin{prop}
\label{prop:opt_router}For a cost budget $B>0$, the optimal router
$r^{*}\colon\mathscr{X}\to\{1,\ldots,M\}$ to the constrained optimization
problem (\ref{eq:optimization}) is 
\begin{align*}
r^{*}(\bx) & =\arg\max_{m\in[M]}\mathbb{E}\left[q(\bx,h^{(m)}(\bx))\mid\bx\right]-\lambda\cdot c^{(m)},
\end{align*}
where the conditional expectation is over the sampled output from the model $h^{(m)}$,
and $\lambda\ge0$ is a Lagrange multiplier inversely proportional
to $B$.
\end{prop}
The result follows from Proposition 1 in \citet{JitNarRaw2025}.
The result states that the choice/model we choose to route a prompt
$\bx$ to is the one that maximizes the average quality, adjusted
additively by the cost of the model. The hyperparameter $\lambda$ 
 controls the trade-off between quality and cost, and is inversely
proportional to the budget $B$. For instance, if $\lambda=0$ (corresponding
to $B=\infty$), then the model with the highest expected quality
for $\bx$ will be chosen, regardless of its cost. Increasing $\lambda$
enforces the routing rule to account more for model costs, in addition
to the expected quality.

\paragraph{Estimating the Optimal Rule}

The optimal rule $r^{*}$ in Proposition \ref{prop:opt_router} depends
on the population conditional expectation  $\gamma^{(m)}(\bx)\defeq\mathbb{E}\left[q(\bx,h^{(m)}(\bx))\mid\bx\right]$,
which is unknown. Following a similar reasoning as in \citet{JitNarRaw2025},
we propose plugging in an empirical estimator $\hat{\gamma}^{(m)}\colon\mathscr{X}\to\mathbb{R}$
in place of $\gamma^{(m)}$,
resulting in the empirical rule $\hat{r}_{\lambda}$:
\begin{align}
\hat{r}_{\lambda}(\bx) & =\arg\max_{m\in[M]}\hat{\gamma}^{(m)}(\bx)-\lambda\cdot c^{(m)}.\label{eq:emp_rule}
\end{align}
For each $m\in[M]$, the idea is to train an estimator $\hat{\gamma}^{(m)}$
to estimate the true expected quality. That is, suppose we are given
a collection of $N$ training prompts $\{\bx_{i}\}_{i=1}^{N}$. For
each prompt $\bx_{i}$, we may sample $S$ times from $h^{(m)}$ to
produce output images $\bi_{i,1}^{(m)}\ldots,\bi_{i,S}^{(m)}$. These
output images allow one to estimate the empirical expectation of
the quality $\hat{y}_{i}\defeq\frac{1}{S}\sum_{s=1}^{S}q(\bx,\bi_{i,s}^{(m)})$.
With the labeled training set $\{(\bx_{i},\hat{y}_{i})\}_{i=1}^{N}$,
we may then proceed to train a predictive model $\hat{\gamma}(\bx)\defeq\left(\hat{\gamma}^{(1)}(\bx),\ldots,\hat{\gamma}^{(M)}(\bx)\right),$
which has $M$ output heads for predicting the expected qualities
of the $M$ models. There are several standard machine learning models
one can use as the model class for $\hat{\gamma}$.

We emphasize that we do not advocate a specific model class as part of our proposal
since different model classes offer distinct properties on training and
inference costs, which may be best tailored to the application. 
What we propose is an application of the generic routing rule in (\ref{eq:emp_rule}) to text-to-image model routing. The rule
is guaranteed to give a good quality-cost trade-off provided that
the estimator $\hat{\gamma}^{(m)}$ well estimates $\gamma^{(m)}$.
In experiments (\Cref{sec:experiments}), we demonstrate estimating
$\gamma^{(m)}$ with two model classes: 1) $K$-nearest neighbors,
and 2) Multi-Layer Perceptron (MLP) with a Transformer backbone \citep{VasShaPar2017}.
Likewise, we do not propose or advocate a specific value of $\lambda$.
The parameter is left to the user as a knob to control the desired
degree of quality-cost trade-off. In experiments, we evaluate our
proposed routing rule by considering a wide range of $\lambda$ and
show the trade-off as a deferral curve (see \Cref{sec:deferral_curve}).
An illustration summarizing our pipeline is displayed in \Cref{fig:pipeline}.

\subsection{Deferral Curve}

\label{sec:deferral_curve}In general, any methods that offer the
ability to trade off quality and cost may be evaluated via a \emph{deferral
curve} \citep{BolWanDek2017,CorDeSMoh2016,GupNarJit2024,NarJitMen2022}.
A deferral curve is a curve showing the average quality against the
average cost, in a quality-cost two-dimensional plane. Specifically,
for our proposed routing rule $\hat{r}_{\lambda}$ in (\ref{eq:emp_rule}),
the curve is precisely given by $\mathscr{C}=\{ (C(\hat{r}_{\lambda}), Q(\hat{r}_{\lambda}))\mid\lambda\in[0,\infty)\}$
where $Q$ and $C$ denote the average quality and cost, and are defined
in Eq. \eqref{eq:avg_qc}. In
practice, the population expectation in $Q$ and $C$ is replaced
with an empirical expectation over examples in a test set. We generate the deferral curve of our routing decision by sweeping $\lambda$ from 0 to a sufficiently large value. This allows the router to transition from maximizing quality at $\lambda=0$ to selecting the cheaper candidates as $\lambda$ increases. More generally,
one evaluates the deferral curve of a method by computing its average
quality and cost as we vary parameters that control the trade-off.
For instance, for the SDXL diffusion model, we may produce a deferral
curve by varying the number of denoising steps. 

\section{Related Work}
\label{sec:related}
\paragraph{Uniform Optimization Strategies for Diffusion Models}
Diffusion models have recently exploded in popularity due to their high performance on tasks such as image and video generation, audio generation, and 3D shape generation \citep{ho2020denoising,ramesh2021zero}.
Latent diffusion models \citep{rombach2022high} have significantly improved training and inference efficiency, but still require a large number of forward denoising neural network evaluations to produce high-quality results. 
To address this, an extensive body of literature has been proposed to optimize and accelerate diffusion models, which are typically applied \emph{uniformly across all prompts}.
For example, optimizing the sampling strategy may enable more efficient denoising computation \citep{li2024snapfusion,chen2023speed,li2023autodiffusion}, such as timestep integration \citep{nichol2021improved} or conditioning on the denoising \citep{preechakul2022diffusion}. Optimizing solvers for the denoising step can also efficiently reduce the computation to avoid re-training or fine-tuning \citep{song2020denoising,lu2022dpm,liu2022pseudo,karras2022elucidating}. Alternatively, reducing the redundant computations by caching the internal results within the denoising network is also explored in \citep{ma2024learning, ma2023deepcache}. Another common approach includes model-based optimizations, such as distilling a fully trained model into a smaller student model that achieves comparable results with fewer denoising steps \citep{sauer2024fast,salimans2022progressive,meng2023distillation,liu2023instaflow} or combining multiple denoising models with different sizes to accelerate the denoising process \citep{yang2023denoising, li2023not, pan2024t}. 
An alternative strategy is to approximate the direct mapping from initial noise to generated images, further reducing the number of denoising steps \citep{luo2023latent,song2023consistency}.

\paragraph{Adaptive Optimization Strategies for Diffusion Models}
Instead of a fixed reduction in computational resources, AdaDiff \citep{tang2023deediff} explores a more dynamic approach where the number of denoising steps is decided based on the uncertainty estimation of the intermediate results during denoising. Our work shares a similar motivation for flexible resource allocation. However, we adaptively allocate resources according to prompt complexity and thus can select the most suitable number of steps or model before any denoising process. 
Concurrently, AdaDiff \citep{Zhang2023AdaDiffAS} tackles optimal number of steps selection using a prompt-specific policy, with a lightweight network trained on a reward function that balances image quality and computational resources. In contrast, we decouple the quality estimation from the routing decision, which allows our framework to adapt to different resource constraints without any retraining.

\paragraph{Learning-To-Defer, and Modeling Routing}
The idea of adaptively invoking a different expert on each input is a widely studied area in machine learning under the topic of \emph{learning to defer}. Here, each expert may be a human expert \citep{MozSon2020,MozLanWei2023,SanErdKon2023}, or a larger model \citep{NarJitMen2022,JitGupMen2023,MaoMohMoh2023,GupNarJit2024}. 
In the latter, depending on the topology or order the models are invoked, a learning-to-defer method may yield a \emph{cascade} if models are arranged in a chain
\citep{WanKonChr2022,JitGupMen2023,KolDenTal2024}; or yield a \emph{routed model} if there is a central routing logic (i.e., the router) which selectively sends input traffic to appropriate models \citep{JiaRenLin2023,MaoMohMoh2023,GupNarJit2024,JitNarRaw2025}.
The latter setup is also known as \emph{model routing} and receives much attention of late, especially in the natural language processing 
 literature. Model routing has been successfully applied to route between many Large Language Model (LLMs) of various sizes and specialties 
 (see \citet{CheZahZou2023,HuBieLi2024,ZhuWuWen2025,OngAlmWu2025,JitNarRaw2025} and references therein).
To our knowledge, our work is one of the first that connects the model routing problem to efficient text-to-image generation.

\section{Experiments}
\label{sec:experiments}
In this section, we show how our proposed routing method (\Cref{sec:main}) can be realized in practice by evaluating its effectiveness on real data.
We experiment with both homogeneous (i.e., all routing candidates are derived from the same diffusion model with different candidate numbers of denoising steps), and heterogeneous settings (i.e.,  the routing candidates also include different generative models).
Our goal is to optimally select the best model (or number of denoising steps) for each input prompt given a specified cost constraint. 

\subsection{Experimental Setup}
\label{sec:exp:setting}
\paragraph{Text-To-Image Generative Models} 
As defined in \Cref{sec:problem}, our method selects from a set of generative models \(\mathscr{H}\) for each input prompt. 
We consider a diverse range of models with varying configurations, each offering a different trade-off between image quality and computational cost: 
\begin{enumerate}[leftmargin=*]
    \item \textsc{SDXL}: a widely-used SD architecture \citep{rombach2022high}. To see the full extent of the achievable trade-off, we consider representative numbers of denoising steps in a wide range between 1 and 100.
    
    \item \textsc{Turbo} \citep{sauer2024fast} and \textsc{Lightning} \citep{Lin2024SDXLLightningPA}: distilled versions of SDXL for faster generation. We use the SDXL variant with 1 step for Turbo, and 4 steps for Lighting.
    
    \item \textsc{DDIM} \citep{song2020denoising}: a non-Markovian diffusion process allowing faster sampling. We use this sampling strategy on the SDXL variant at 50 steps.
    
    \item \textsc{DeepCache} \citep{ma2023deepcache}: a caching method that reduces redundant computation in SDXL. 
    We use the official implementation released from \citet{ma2023deepcache}, and set the cache interval parameter to 3.
    
    \item \textsc{Infinity} \citep{han2024infinityscalingbitwiseautoregressive}: a non-diffusion, auto-regressive text-to-image model based on the Transformer encoder-decoder. We use the pre-trained Infinity-2B variant with a visual vocabulary size of $2^{32}$.
\end{enumerate}

\paragraph{Quality Metrics} 
The effectiveness of generative models largely depends on the criteria used to evaluate their output. Our proposed method can adaptively identify the optimal allocation of generative model for \emph{any} instance-level image quality metric. As there is no consensus on the optimal metric for evaluating image quality, we explore several widely-used metrics: \textit{CLIPScore} \citep{radford2021learning} for text-image semantic alignment, \textit{ImageReward} \citep{xu2023imagereward} with a reward model tuned to human preferences, and \textit{Aesthetic Score} \citep{laion_aesthetics_predictor_v1} trained on human aesthetic ratings from LAION \citep{laion_dataset}. Additionally, we also introduce \textit{Sharpness} metric adapted from \citet{paris2011local}, defined as, $q_{\mathrm{Sharp}}(\bx,\bi)=\frac{\sum_{ij}\left(\bi_{ij}-[\bi\circledast G]_{ij}\right)^{2}}{\sum_{ij}\bi_{ij}^{2}},$ where $\circledast$ denotes the convolution operator,  $\bi_{i,j}$ is the pixel intensity at location $(i, j)$, and $G$ is a Gaussian kernel with standard deviation of 1. Intuitively, this metric measures the relative distance between the given image $\bi$ and itself after a Gaussian blur filter is applied. 

\paragraph{Quality Estimator $\hat{\gamma}$}
One of the key components of our routing method is the quality estimator which estimates the expected quality of the $m$-th model given an input prompt (see $\hat{\gamma}^{m}$ in Eq. \eqref{eq:emp_rule}).
We explore two  model classes: a \textsc{K-Nearest Neighbors ($K$-NN)} model and a \textsc{Transformer}-based model. 
Both of these models incur a negligible inference cost: less than 0.001 TFLOPs compared to $1.5$ TFLOPs of the smallest base model in the pool  (Infinity).

The \textsc{$K$-NN} approach provides a non-parametric way to estimate quality by averaging the quality scores of $K$ nearest training prompts in the space of CLIP embeddings \citep{radford2021learning}.
This method is simple, and can generalize well with sufficient data. 
%
The Transformer model takes as input the per-token embeddings produced by the frozen CLIP text encoder. A two-layer   MLP with $M$ output heads is added to each output token embedding. Pooling across all tokens gives $M$ output scores $\hat{\gamma}^{(1)}(\bx), \ldots, \hat{\gamma}^{(M)}(\bx)$ (see Eq. \eqref{eq:emp_rule}), each estimating the expected quality of the $m$-th model on prompt $\bx$ (see Appendix \Cref{sec:supp:model_architect} for details).

All base models except Infinity already use CLIP embeddings, making router overhead negligible. Infinity uses Flan-T5 embeddings ($\approx13$ GFLOPs overhead), but this cost is minimal compared to one SDXL call ($\approx200$ TFLOPs for 17 steps).

We train a separate model for each of the quality metrics considered. 
In each case, the quality scores are linearly scaled across all training examples to be in [0, 1]. 
These scaled metrics are treated as ground-truth probabilities, and the model is trained by minimizing the sum of the sigmoid cross-entropy losses across all heads. 


\subsection{Dataset Details} 
\label{sec:method:data}
We utilize two datasets: 1) the COCO captioning dataset \citep{lin2014microsoft}, which contains high-quality and detailed image captioning, and 2) the DiffusionDB dataset \citep{wangDiffusionDBLargescalePrompt2022}, which contains a larger collection of realistic, user-generated text prompts for text-to-image generation. From both datasets, we sub-sample prompts by retaining only those with pairwise CLIP similarity below 0.75, resulting in a diverse set of 18,384 prompts in COCO dataset, and 97,841 prompts on the DiffusionDB dataset. We split each dataset independently into 80\% for training, 10\% for validation, and 10\% for testing. We then generate images from those prompts using all the base text-to-image models as described earlier. 
For SDXL, we generate images with various numbers of denoising steps ranging from 1 to 100. 
The costs in terms of FLOPs from these candidates cover the full range of costs of all other baselines.

For each model, we generate four images per prompt (i.e., $S=4$ in \Cref{sec:opt_rule}) using different random seeds, with a fixed seed across different numbers of steps for SDXL. 
The generated images for each prompt $\bx_i$ allow us to compute the average quality metric, which is then used as the training label $\hat{y}_i$ (as described in \Cref{sec:opt_rule}). 
Unless otherwise specified, we use the widely used Euler Scheduler \citep{karras2022elucidating} for diffusion-based image generation.



\newcommand{\includelegend}[1]{%
\includegraphics[trim={15.5cm 0 0 0},clip,width=#1]{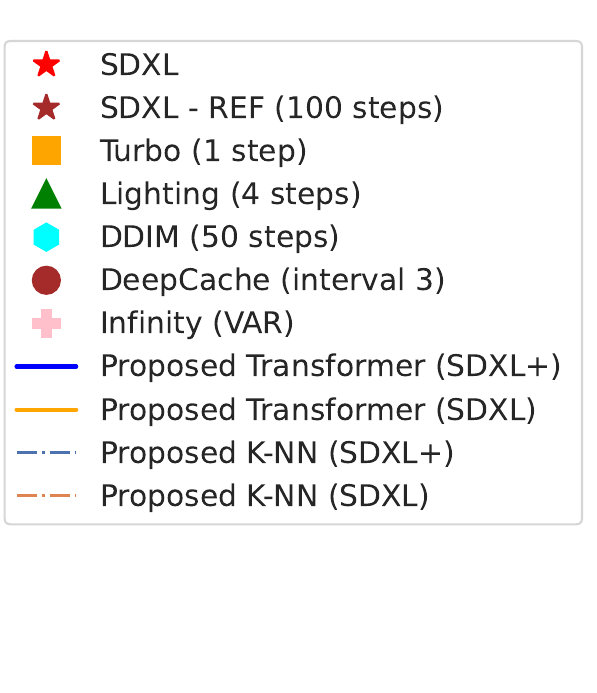}
}
\begin{figure}
  \begin{minipage}[b]{\linewidth}
    \begin{subfigure}{0.34\textwidth}
        \begin{tikzpicture}
        \node (img) at (0,0) {\includegraphics[trim={0 0 10.2cm 0},clip,width=\textwidth]{assets/iccv/deferral_curve_CLIP.pdf}};
        
        \node[scale=0.7,text width=4cm] (l) at (1.3, -0.92) {%
        \footnotesize \color{black!70!white}
            Lighting: $(33.9, 0.2749)$\\
            Infinity: \,\, $(1.5, 0.2672)$
            };
        \end{tikzpicture}
        \caption{CLIPScore}
        \label{fig:deferral_curve_clip}
    \end{subfigure}
    \hspace{.5mm}
    \begin{subfigure}{0.35\textwidth}
        \includegraphics[trim={0 0 10.2cm 0},clip,width=\textwidth]{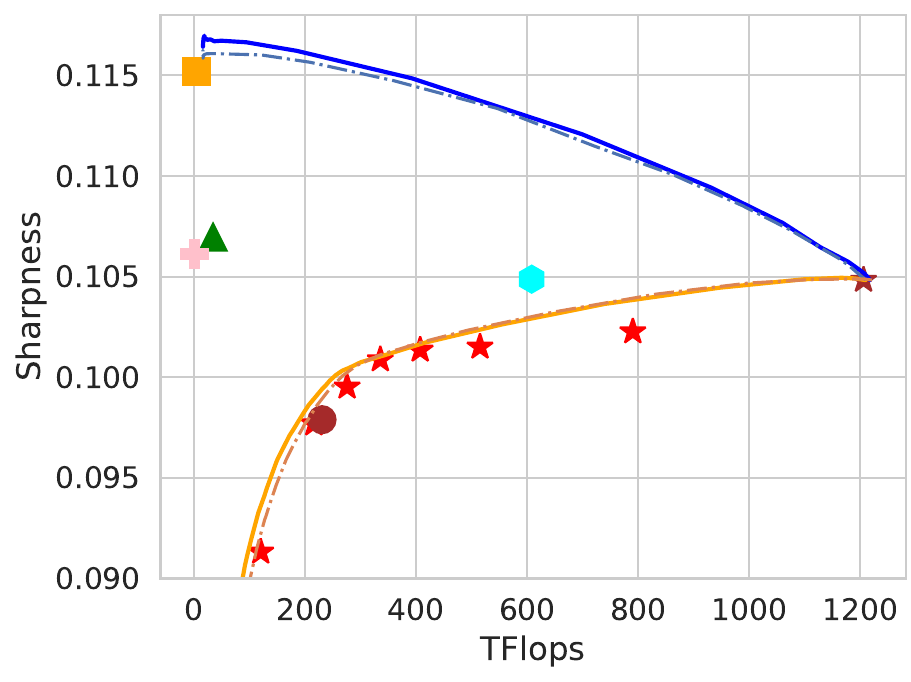}
        \caption{Sharpness}
        \label{fig:deferral_curve_sharpness}
    \end{subfigure}
     \begin{subfigure}[t]{0.29\textwidth}
        \includelegend{0.88\textwidth}
        \vspace*{-3mm}
    \end{subfigure}
    %
  \end{minipage}
  \caption{Deferral curves of our proposed methods and baselines on the COCO dataset. Each data point represents the average quality and cost across the entire test set ($1839$ prompts). Quality is measured by CLIPScore in (a) and pixel sharpness in (b), as defined in \Cref{sec:exp:setting}. 
  Our Proposed Transformer (SDXL+), which selects from all available SDXL denoising steps and baseline models, offers the best quality-cost trade-off (TFLOPs). Baselines that are not visible are shown at the bottom-right corner in the format of (cost, CLIPScore).
    } 
\hspace{1mm}
\vspace{-5mm}
\label{fig:main_plot}
\end{figure}

\subsection{Experiments on COCO dataset}
\label{sec:coco}
We present experimental results on a subset of COCO's test set \citep{lin2014microsoft} consisting of 1.8k image-caption pairs in \Cref{fig:main_plot}.
We evaluate the deferral curves (see \Cref{sec:deferral_curve}) of our proposed method and all the baselines.
The results are shown in \Cref{fig:deferral_curve_clip,fig:deferral_curve_sharpness} for the two different quality metrics: CLIPScore and image sharpness (\Cref{sec:exp:setting}), respectively.
The deferral curves plot average quality against average cost measured in TFLOPs (Tera Floating Point Operations).
Baselines that do not support dynamic quality-cost trade-off are shown as isolated dots in the same quality-cost plane; these baselines use the same compute cost for image generation for each input prompt.
For instance, each point \textcolor{red}{$\bigstar$} of SDXL represents the performance of the SDXL model with the number of denoising steps fixed. For our proposed methods,
\textit{Proposed (SDXL)} refers to the homogeneous configuration in which the model candidate set $\mathscr{H}$ consists solely of the SDXL model at multiple numbers of denoising steps settings. 
\textit{Proposed (SDXL+)} extends this configuration by incorporating other text-to-image models considered, namely, \textsc{Turbo}, \textsc{DDIM}, \textsc{DeepCache}, and \textsc{Infinity}. Each of these has two variants based on Transformer or $K$-NN as the model class for estimating the expected quality metric.

\textbf{Homogeneous vs. Heterogeneous setting}
In both settings, our methods outperform baselines with static inference costs per prompt. The heterogeneous setting further benefits from models with strong quality-to-cost trade-offs (e.g., \textsc{Infinity}, \textsc{Turbo}), improving our dynamic routing's effectiveness and cost-efficiency. Moreover, our strategy remains adaptive, seamlessly allocating prompts to higher-performance models when additional computational resources are available, improving performance beyond what is attainable using each model alone (see Appendix \Cref{sec:supp:selection_rates} for details on model selection rates).

\textbf{Transformer vs. KNN}
 Between the two proposed variants, the Transformer-based variant generally outperforms the $K$-NN variant, suggesting that directly learning to predict the quality metric can be more effective than estimating it from neighboring prompts.

\begin{table}[t]
    \centering
    \resizebox{\textwidth}{!}{%
    \begin{tabular}{lcccccccc}
    \toprule
     \multicolumn{6}{l}{\textbf{CLIPScore} {\color{teal}$\uparrow$}}\\
     \midrule
                   & Infinity & Turbo  & Lighting & SDXL-9  & DeepCache & SDXL-22 & DDIM   & SDXL-65 \\ 
    \midrule
    Ours (S=1)     & 0.2816   & 0.2814 & 0.2828   & 0.2828  & 0.2828    & 0.2828  & 0.2824 & 0.2819  \\
    Ours (S=3)     & 0.2816   & 0.2816 & 0.2831   & \bfseries{0.2832}  & \bfseries{0.2832}    & \bfseries{0.2832}  & 0.2828 & 0.2819  \\
    Fixed          & 0.2816   & 0.2794 & 0.2798   & 0.2773  & 0.2749    & 0.2805  & 0.2820 & 0.2819  \\
    \midrule \midrule
    Win Rate (S=1) & -        & 1      & 1        & 1       & 1         & 1       & 0.98   & -       \\
    Win Rate (S=3) & -        & 1      & 1        & 1       & 1         & 1       & 1      & -       \\ 
    \midrule
    \rowcolor{gray!15}
    TFLOPs         & 1.5      & 1.54   & 23.92    & 107.73  & 210       & 263.34  & 598.5  & 778.05  \\ 
    \bottomrule
    \end{tabular}%
    }
    \caption{We train the k-NN quality estimator over $T=100$ trials on COCO. In each trial, we generate $S$ images for each training prompt with 1) $S = 1$ and 2) $S=3$. Each of these two cases results in $T$ k-NN models, and hence $T$ (random) deferral curves evaluated on the same test set as used in Fig 3. We report the mean performance (CLIPScore) across the $T$ trials for our approach. }
    \label{tab:winrate_var}
\end{table}

\subsection{Effect of Sample Size S}
As the estimation error depends on the randomness in image generation, we aggregate signals from multiple generated images by averaging their quality metric values across multiple random seeds. To further quantify variability across trials, we vary the number of generated images for each prompt and train a separate model for each case. In~\Cref{tab:winrate_var}, we report the mean performance of CLIPScores on the COCO dataset, as well as the Win Rate, defined as the fraction of trials that our router has higher average quality across the test set than the baseline. The maximum standard deviation across trials of both above variants is less than $2 \times 10^{-4}$ throughout the cost range. 

The results show that using S=3 improves the test performance compared to S=1. 
Additionally, the win rate of “Ours (S=1)” compared to the baseline is 100\% in almost all cost ranges. This win rate implies that our approaches are statistically significantly better than the baseline according to the sign test at significance level $a<10^{-6}$. This means using one image per prompt is already sufficient to improve the baseline of using fixed compute costs. Deviation across trials is minimal relative to the mean metric, suggesting that S=3 is sufficient. In all other experiments in the paper, we used S=4.

\begin{table}[t]
  \centering
  \caption{Quality-cost trade-off of our proposed approach on DiffusionDB (\Cref{sec:diffusiondb}). 
  We report the average quality achieved by our routing approach when operating at the cost 
  (TFLOPs) of each model in the pool. 
  For each metric, the highest score achieved between our approach and the fixed baseline is highlighted in bold. The Oracle performance is provided in gray for context.
  }
  \label{tab:quantiative_results}
  \resizebox{\linewidth}{!}{
  \begin{tabular}{l*{9}{r}}
    \toprule
    \addlinespace[0.5em]
    \multicolumn{10}{l}{\textbf{CLIPScore} \citep{radford2021learning} \text{ } {\color{teal}$\uparrow$}}\\
    \midrule
    \addlinespace[0.2em]
    {\color{gray} Oracle}
       & {\color{gray} $0.259_{\,\pm\,6\text{e-4}}$} & {\color{gray} $0.309_{\,\pm\,4\text{e-4}}$} & {\color{gray} $0.323_{\,\pm\,4\text{e-4}}$} & {\color{gray} $0.334_{\,\pm\,4\text{e-4}}$} & {\color{gray} $0.337_{\,\pm\,4\text{e-4}}$} 
       & {\color{gray} $0.337_{\,\pm\,4\text{e-4}}$} & {\color{gray} $0.336_{\,\pm\,4\text{e-4}}$} & {\color{gray} $0.336_{\,\pm\,4\text{e-4}}$} & {\color{gray} $0.318_{\,\pm\,4\text{e-4}}$} \\
    Ours  
       &  $0.259_{\,\pm\,6\text{e-4}}$  &  $0.304_{\,\pm\,4\text{e-4}}$  &  $0.308_{\,\pm\,4\text{e-4}}$  &  $0.314_{\,\pm\,4\text{e-4}}$  &  $0.316_{\,\pm\,4\text{e-4}}$ 
       &  $0.317_{\,\pm\,4\text{e-4}}$  &  $\textbf{0.318}_{\,\pm\,4\text{e-4}}$  &  $0.318_{\,\pm\,4\text{e-4}}$  &  $0.318_{\,\pm\,4\text{e-4}}$ \\
    Fixed 
       &  $0.259_{\,\pm\,6\text{e-4}}$  &  $0.304_{\,\pm\,4\text{e-4}}$  &  $0.300_{\,\pm\,4\text{e-4}}$  &  $0.308_{\,\pm\,4\text{e-4}}$  &  $0.316_{\,\pm\,4\text{e-4}}$ 
       &  $0.315_{\,\pm\,4\text{e-4}}$  &  $0.317_{\,\pm\,4\text{e-4}}$  &  $0.315_{\,\pm\,4\text{e-4}}$  &  $0.318_{\,\pm\,4\text{e-4}}$ \\
    \midrule
    \multicolumn{10}{l}{\textbf{Sharpness} (\Cref{sec:exp:setting}) \text{ } {\color{teal}$\uparrow$}}\\
    \midrule
    \addlinespace[0.2em]
    {\color{gray} Oracle}
       & {\color{gray} $0.131_{\,\pm\,4\text{e-4}}$} & {\color{gray} $0.140_{\,\pm\,3\text{e-4}}$} & {\color{gray} $0.144_{\,\pm\,3\text{e-4}}$} & {\color{gray} $0.145_{\,\pm\,3\text{e-4}}$} & {\color{gray} $0.146_{\,\pm\,3\text{e-4}}$} 
       & {\color{gray} $0.146_{\,\pm\,3\text{e-4}}$} & {\color{gray} $0.145_{\,\pm\,4\text{e-4}}$} & {\color{gray} $0.145_{\,\pm\,4\text{e-4}}$} & {\color{gray} $0.126_{\,\pm\,3\text{e-4}}$} \\
    Ours  
       &  $0.131_{\,\pm\,4\text{e-4}}$  &  $0.135_{\,\pm\,3\text{e-4}}$  &  $0.135_{\,\pm\,3\text{e-4}}$  &  $0.136_{\,\pm\,3\text{e-4}}$  &  $\textbf{0.137}_{\,\pm\,3\text{e-4}}$ 
       &  $0.137_{\,\pm\,3\text{e-4}}$  &  $0.137_{\,\pm\,3\text{e-4}}$  &  $0.137_{\,\pm\,3\text{e-4}}$  &  $0.126_{\,\pm\,3\text{e-4}}$ \\
    Fixed 
       &  $0.131_{\,\pm\,4\text{e-4}}$  &  $0.122_{\,\pm\,3\text{e-4}}$  &  $0.110_{\,\pm\,3\text{e-4}}$  &  $0.103_{\,\pm\,2\text{e-4}}$  &  $0.101_{\,\pm\,2\text{e-4}}$ 
       &  $0.114_{\,\pm\,2\text{e-4}}$  &  $0.123_{\,\pm\,3\text{e-4}}$  &  $0.107_{\,\pm\,3\text{e-4}}$  &  $0.126_{\,\pm\,3\text{e-4}}$ \\

    \midrule
    \multicolumn{10}{l}{\textbf{Aesthetic Score} \citep{laion_aesthetics_predictor_v1} \text{ } {\color{teal}$\uparrow$}}\\
    \midrule
    \addlinespace[0.2em]
    {\color{gray} Oracle}
       & {\color{gray} $6.824_{\,\pm\,8\text{e-3}}$} & {\color{gray} $7.132_{\,\pm\,8\text{e-3}}$} & {\color{gray} $7.385_{\,\pm\,8\text{e-3}}$} & {\color{gray} $7.426_{\,\pm\,7\text{e-3}}$} & {\color{gray} $7.421_{\,\pm\,8\text{e-3}}$} 
       & {\color{gray} $7.416_{\,\pm\,8\text{e-3}}$} & {\color{gray} $7.284_{\,\pm\,8\text{e-3}}$} & {\color{gray} $7.284_{\,\pm\,8\text{e-3}}$} & {\color{gray} $6.707_{\,\pm\,8\text{e-3}}$} \\
   Ours  
       &  $6.824_{\,\pm\,8\text{e-3}}$  &  $6.913_{\,\pm\,9\text{e-3}}$  &  $\textbf{7.042}_{\,\pm\,9\text{e-3}}$  &  $7.032_{\,\pm\,9\text{e-3}}$  &  $7.012_{\,\pm\,9\text{e-3}}$ 
       &  $7.012_{\,\pm\,9\text{e-3}}$  &  $6.935_{\,\pm\,9\text{e-3}}$  &  $6.935_{\,\pm\,9\text{e-3}}$  &  $6.707_{\,\pm\,8\text{e-3}}$ \\
    Fixed 
       &  $6.824_{\,\pm\,8\text{e-3}}$  &  $6.780_{\,\pm\,9\text{e-3}}$  &  $7.010_{\,\pm\,9\text{e-3}}$  &  $6.285_{\,\pm\,8\text{e-3}}$  &  $6.625_{\,\pm\,9\text{e-3}}$ 
       &  $6.588_{\,\pm\,8\text{e-3}}$  &  $6.690_{\,\pm\,8\text{e-3}}$  &  $6.600_{\,\pm\,9\text{e-3}}$  &  $6.707_{\,\pm\,8\text{e-3}}$ \\

    \midrule
    \multicolumn{10}{l}{\textbf{ImageReward} \citep{xu2023imagereward} \text{ } {\color{teal}$\uparrow$}}\\
    \midrule
    \addlinespace[0.2em]
    {\color{gray} Oracle}
       & {\color{gray} $1.029_{\,\pm\,9\text{e-3}}$} & {\color{gray} $1.303_{\,\pm\,6\text{e-3}}$} & {\color{gray} $1.416_{\,\pm\,5\text{e-3}}$} & {\color{gray} $1.446_{\,\pm\,5\text{e-3}}$} & {\color{gray} $1.446_{\,\pm\,5\text{e-3}}$} 
       & {\color{gray} $1.444_{\,\pm\,5\text{e-3}}$} & {\color{gray} $1.376_{\,\pm\,5\text{e-3}}$} & {\color{gray} $1.376_{\,\pm\,5\text{e-3}}$} & {\color{gray} $0.891_{\,\pm\,7\text{e-3}}$} \\
   Ours  
       &  $1.029_{\,\pm\,9\text{e-3}}$  &  $1.083_{\,\pm\,8\text{e-3}}$  &  $\textbf{1.086}_{\,\pm\,8\text{e-3}}$  &  $1.086_{\,\pm\,8\text{e-3}}$  &  $1.079_{\,\pm\,8\text{e-3}}$ 
       &  $1.076_{\,\pm\,8\text{e-3}}$  &  $1.037_{\,\pm\,8\text{e-3}}$  &  $1.037_{\,\pm\,8\text{e-3}}$  &  $0.891_{\,\pm\,7\text{e-3}}$ \\
    Fixed 
       &  $1.029_{\,\pm\,9\text{e-3}}$  &  $0.960_{\,\pm\,8\text{e-3}}$  &  $0.932_{\,\pm\,8\text{e-3}}$  &  $0.497_{\,\pm\,8\text{e-3}}$  &  $0.809_{\,\pm\,9\text{e-3}}$ 
       &  $0.769_{\,\pm\,8\text{e-3}}$  &  $0.866_{\,\pm\,8\text{e-3}}$  &  $0.861_{\,\pm\,9\text{e-3}}$  &  $0.891_{\,\pm\,7\text{e-3}}$ \\
    \addlinespace[0.2em]
    \midrule
    \midrule
    \rowcolor{gray!15}
       &  \textbf{INFI}  &  \textbf{TURB}  &  \textbf{LIGH}  &  \textbf{SDXL}  &  \textbf{DEEP}  &  \textbf{SDXL}  &  \textbf{SDXL}  &  \textbf{DDIM}  &  \textbf{SDXL} \\
    \rowcolor{gray!15}
    \textbf{TFLOPs}
       &  {\bfseries 1.50}  &  {\bfseries 1.54}  &  {\bfseries 23.92}
       &  {\bfseries 119.70}  &  {\bfseries 210.00}  &  {\bfseries 239.40}
       &  {\bfseries 598.50}  &  {\bfseries 598.50}  &  {\bfseries 1197.00} \\
    \bottomrule
  \end{tabular}
  }
\end{table}

\subsection{Experiments on DiffusionDB dataset}
\label{sec:diffusiondb}
In this section, we present results on a subset of prompts from the DiffusionDB dataset \citep{wangDiffusionDBLargescalePrompt2022}, which aligns more closely with real-world prompts used in text-to-image generation. We evaluate the performance across four metrics: \textit{CLIPScore}, \textit{ImageReward}, \textit{Aesthetic Score}, and \textit{Sharpness}.

Quantitative results comparing our dynamic routing method to the fixed-model baselines are summarized in \Cref{tab:quantiative_results}. This table effectively captures the trade-offs shown in the deferral curves at a specific cost equal to each baseline. We use KNN as a quality estimator to efficiently evaluate multiple metrics at scale. For reference, we also provide the Oracle performance, which selects the optimal candidate for each prompt based on ground-truth quality scores rather than predicted estimates. This is an absolute upper bound on the best quality-cost trade-off attainable by any routing methods. In practice, it is extremely challenging to realize a quality-cost operating point that is close to that of the Oracle (see \cite{HuBieLi2024}). The results show that our method consistently matches or exceeds fixed-model baseline performance across all four quality metrics. 
Additionally, the highest value of each score (highlighted in \Cref{tab:quantiative_results} in bold) is attainable \emph{only} with our routing strategy. In other words, even under an unconstrainedand computational budget, none of the individual baselines can attain the quality that our adaptive routing achieves through prompt-based allocation across the model pool. 

\begin{wraptable}{r}{0.3\textwidth}
\vspace{-2mm} 
\caption{Cost ratio (\%) of our method compared to baselines to match the quality score (Sharpness) }
\label{tab:cost_ratio}
\begin{tabular}{lcc}
\toprule
\textbf{Model}  &  \textbf{Our cost} \\
\midrule
Infinity     &  100\% \\
Turbo        &  97.40\% \\
Lighting   	  &  6.27\%   \\
DeepCache    &  0.71\%  \\
DDIM         &  0.25\%  \\
SDXL100     &  0.13\%   \\
\bottomrule
\end{tabular}
\vspace*{-8mm}
\end{wraptable}

\Cref{tab:cost_ratio} quantifies the computational cost reduction achieved by our routing method compared to the baseline at equivalent quality levels (on Sharpness metric). For inherently efficient models (e.g. Infinity\citep{han2024infinityscalingbitwiseautoregressive}, Turbo \citep{sauer2024fast}), the savings appear marginal. However, compared to Lighting \citep{Lin2024SDXLLightningPA}, a \emph{distilled} SDXL variant, our method achieves the same performance at only 6\% of its computational cost.
For higher-performance models, such as SDXL at 100 denoising steps, the savings are even more significant.

\subsection{Incorporating FLUX.1-dev model}
The quality-cost trade-off of our router improves as we add better models to the pool. With the same experimental setup on DiffusionDB as in \Cref{sec:diffusiondb}, we now add a high-performing model FLUX.1-dev\footnote{\url{https://huggingface.co/black-forest-labs/FLUX.1-dev}} at three different denoising steps (1, 15, and 30), resulting in a total of 12 models in the routing pool. 
The addition of a new model results in an improved performance across all cost ranges compared to using any individual model alone, as shown in \Cref{tab:add_flux}.  Specifically, at an average cost of less than 600 TFLOPs, our router achieves an aesthetic score of 7.109, surpassing the best single model (FLUX-30) which scores 6.98 at a fixed cost of 1785.6 TFLOPs. The router learns that the most powerful model (e.g., FLUX-30) is not necessarily the best for every input; a cheaper model can sometimes yield superior results on prompts for which it is better suited.

\begin{table} 
    \centering
    \caption{Average quality achieved by our routing approach when operating at the cost (TFLOPs) of each model in the pool. In this experiment, we include FLUX.1-dev at 1, 15, and 30 denoising steps, for the total of 12 models in the candidate pool.}
    \setlength{\tabcolsep}{3pt} 
    \resizebox{\textwidth}{!}{%
    \begin{tabular}{lcccccccccccc}
    \toprule
     \multicolumn{13}{l}{\textbf{Aesthetic Score} {\color{teal}$\uparrow$}}\\
     \midrule
     & Infinity & Turbo & Lighting & FLUX-1 & SDXL-10 & DeepCache & SDXL-20 & DDIM & SDXL-50 & FLUX-15 & SDXL100 & FLUX-30 \\ 
    \midrule
    Ours & 6.824 & 6.913 & 7.045 & 7.055 & 7.068 & 7.082 & 7.087 & \textbf{7.109} & 7.109 & 7.108 & 7.090 & 6.983 \\
    Fixed & 6.824 & 6.780 & 7.010 & 3.160 & 6.285 & 6.625 & 6.588 & 6.600 & 6.690 & 6.952 & 6.707 & 6.983 \\ 
    \midrule
    \rowcolor{gray!15}
    TFLOPs & 1.50 & 1.54 & 23.92 & 59.52 & 119.70 & 210 & 239.4 & 598.5 & 598.5 & 892.8 & 1197.00 & 1785.60 \\
    \rowcolor{gray!15}
    Fraction of TFLOPs & 0.1\% & 0.1\% & 1.3\% & 3.3\% & 6.7\% & 11.8\% & 13.4\% & 33.5\% & 33.5\% & 50.0\% & 67.0\% & 100.0\% \\ 
    \bottomrule
    \end{tabular}%
    }
    \label{tab:add_flux}
\end{table}


\subsection{Qualitative Comparison}
\begin{figure}[t]
\centering
\includegraphics[width=\linewidth]{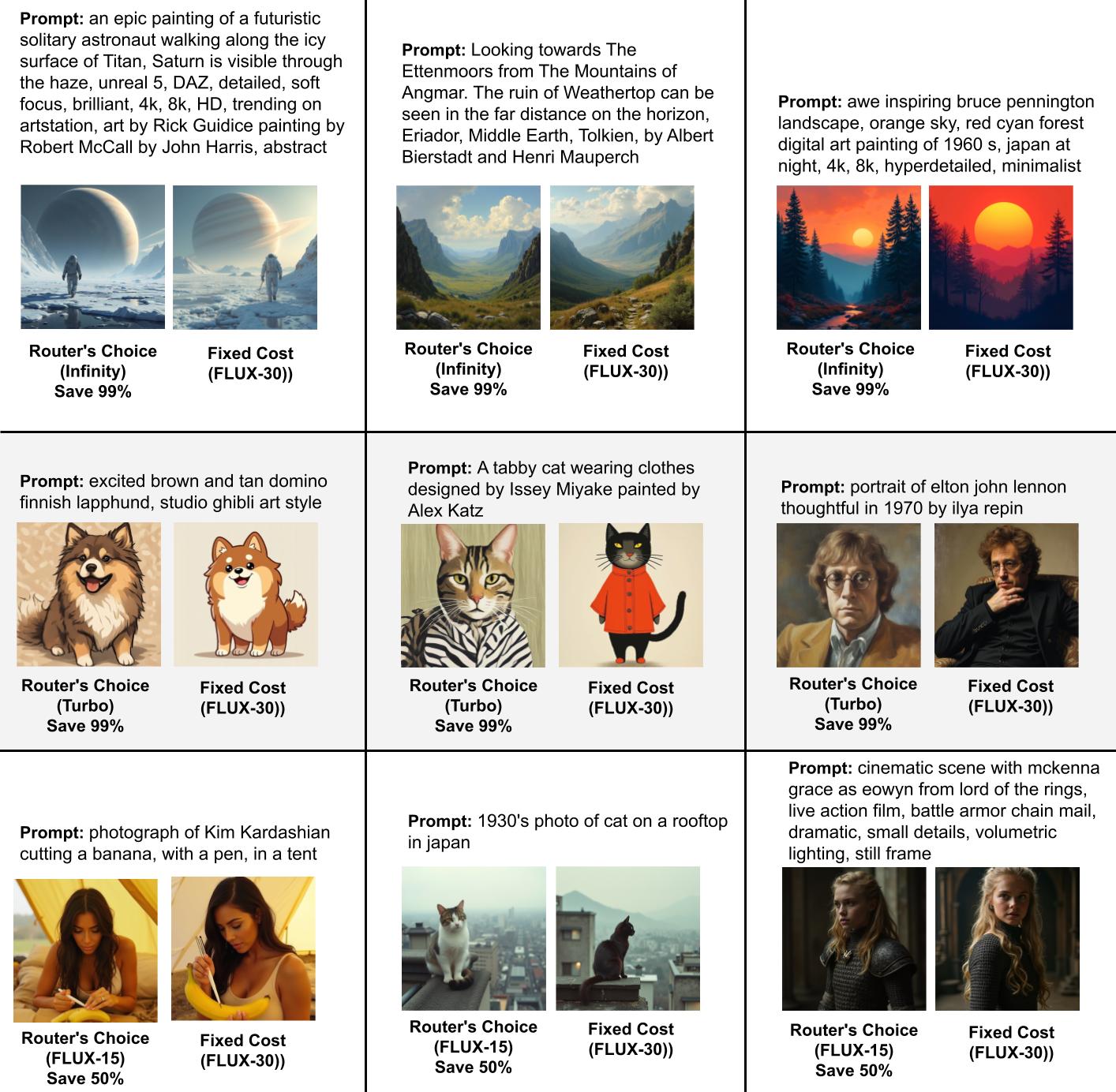}
\caption{Example results from our routing framework compared to a fixed FLUX model (30 denoising steps). Examples are grouped by the three most frequently selected models: Infinity, Turbo, and FLUX-15. We observe a tendency for the router to select Infinity for scenic prompts, while Turbo is often preferred for cartoon and painting styles. FLUX-15 (FLUX at 15 steps) appears to be favored for general prompts, especially those involving humans.
}
\label{fig:qual_comparison}
\vspace{-3mm}
\end{figure}

To better illustrate the effects of routing, we further present qualitative comparisons between our routing results and the FLUX model (with 30 denoising steps) in~\Cref{fig:qual_comparison}. We perform routing using Aesthetic Score, which has been shown to correlate well with human preference (\Cref{sec:hpsv2}). Note that in this case, our routing framework can be used to select the number of denoising steps (i.e. the same FLUX model with 15 denoising steps) or to select an entirely different model (Infinity, Turbo, etc.). Interestingly, we observe that the router tends to select \emph{Infinity} for scenic prompts and \emph{Turbo} for stylized (cartoon/painting) images, whereas for general/human related prompts it typically reduces the number of denoising steps from 30 to 15.

\section{Conclusion and Future Work}
\label{sec:limitation}
In this paper, we present \methodName, a cost-aware routing approach that dynamically selects optimal models and numbers of denoising steps based on prompt complexity. We show that incorporating multiple base models, such as distilled versions of diffusion models and alternative architectures, improves the quality–cost trade-off. Extensive experiments on COCO and DiffusionDB datasets across multiple quality metrics validate our method's effectiveness and generalization capability. Nevertheless, several limitations are worth highlighting. To determine optimal routing decisions, \methodName relies on estimating the expected quality \emph{per} prompt, which excludes metrics such as Fr\'{e}chet Inception Distance (FID) \citep{heusel2017gans} that measure statistical similarity across the entire image distribution. Additionally, the routing problem we consider assumes a static pool setup where the set of text-to-image models is considered fixed. In practice, the model pool can change which can add additional maintenance cost of retraining the router.
Addressing these limitations remains a direction for future research.

%
%

\bibliographystyle{tmlr}

\bibliography{ref}


\clearpage
\appendix





{
\centering
\Large
\textbf{\ourtitle}\\
\vspace{0.5em}Appendix \\
\vspace{1.0em}
}



\section{Human Preference Score}
\label{sec:hpsv2}
To quantitatively evaluate how our routing decisions from each metric align with human perception, we adopt the Human Preference Score v2 (HPSv2) benchmark \citep{wu2023human2}. We trained four separate routers, one for each metric (CLIPScore, Sharpness, Aesthetic Score, and ImageReward), and then report the average HPSv2 quality score achieved by our routing approach when operating at the cost (TFLOPs) of each model in the pool. For this analysis, we also expand our candidate pool to include the recent, state-of-the-art FLUX model (FLUX.1-dev with 30 denoising steps)\footnote{\url{https://huggingface.co/black-forest-labs/FLUX.1-dev}}. The results are shown below.

\begin{center}
  \centering
    \label{tab:hpsv2}
    \resizebox{\linewidth}{!}{
    \begin{tabular}{l*{10}{r}}
        \toprule
        \addlinespace[0.5em]
        \multicolumn{11}{l}{\textbf{HPSv2 score} {\color{teal}$\uparrow$}}\\
        \midrule
        \addlinespace[0.2em]
         & Infinity & Turbo & Lighting & SDXL-10 & DeepCache & SDXL-20 & DDIM & SDXL-50 & SDXL100 & FLUX \\ 
        Fixed Baseline  &  0.293  &  0.288  &  0.304  &  0.251  &  0.277  &  0.275  &  0.284  &  0.284  &  0.286  &  0.298\\
        \midrule
        Ours (CLIPScore) &  0.293  &  0.288  &  0.287  &  0.282  &  0.277  &  0.278  &  0.284  &  0.284  &  0.286  &  0.298\\
        Ours (Sharpness) &  0.293  &  0.293  &  0.293  &  0.293  &  0.293  &  0.293  &  0.295  &  0.295  &  0.296  &  0.298\\
        Ours (ImageReward) &  0.293  &  0.293  &  0.293  &  0.293  &  0.293  &  0.293  &  0.295  &  0.295  &  0.296  &  0.298\\
        Ours (Aesthetic) &  0.293  &  0.293  &  0.304  &  0.304  &  0.303  &  0.303  &  0.302  &  0.302  &  0.300  &  0.298\\
    
     \bottomrule
        \rowcolor{gray!15}
        \textbf{TFLOPs}
         &  { 1.50}  &  { 1.54}  &  { 23.92}
         &  { 119.70}  &  { 210.00}  &  { 239.40}
         &  { 598.50}  &  { 598.50}  &  { 1197.00} &  { 1785.60}\\
         \bottomrule
    \end{tabular}
    }
\end{center}
Notably, the results indicate that routing with the Aesthetic Score produces outcomes that correlate most closely with human preferences as evaluated by HPSv2. This is likely because the Aesthetic Score model itself is trained on human ratings of aesthetic value. Note that when the cost constraint is set to match the cheapest (Infinity) or most expensive (FLUX) model, the router must select that model for all prompts, resulting in identical performance.
    
\section{Additional Qualitative Analysis}
\begin{figure}[t]
\centering
\begin{minipage}[b]{.48\textwidth}
\textbf{Success case}. 
A young kid stands before a birthday cake decorated with Captain America \\
\begin{subfigure}[t]{0.49\textwidth}
    \includegraphics[width=0.92\linewidth]{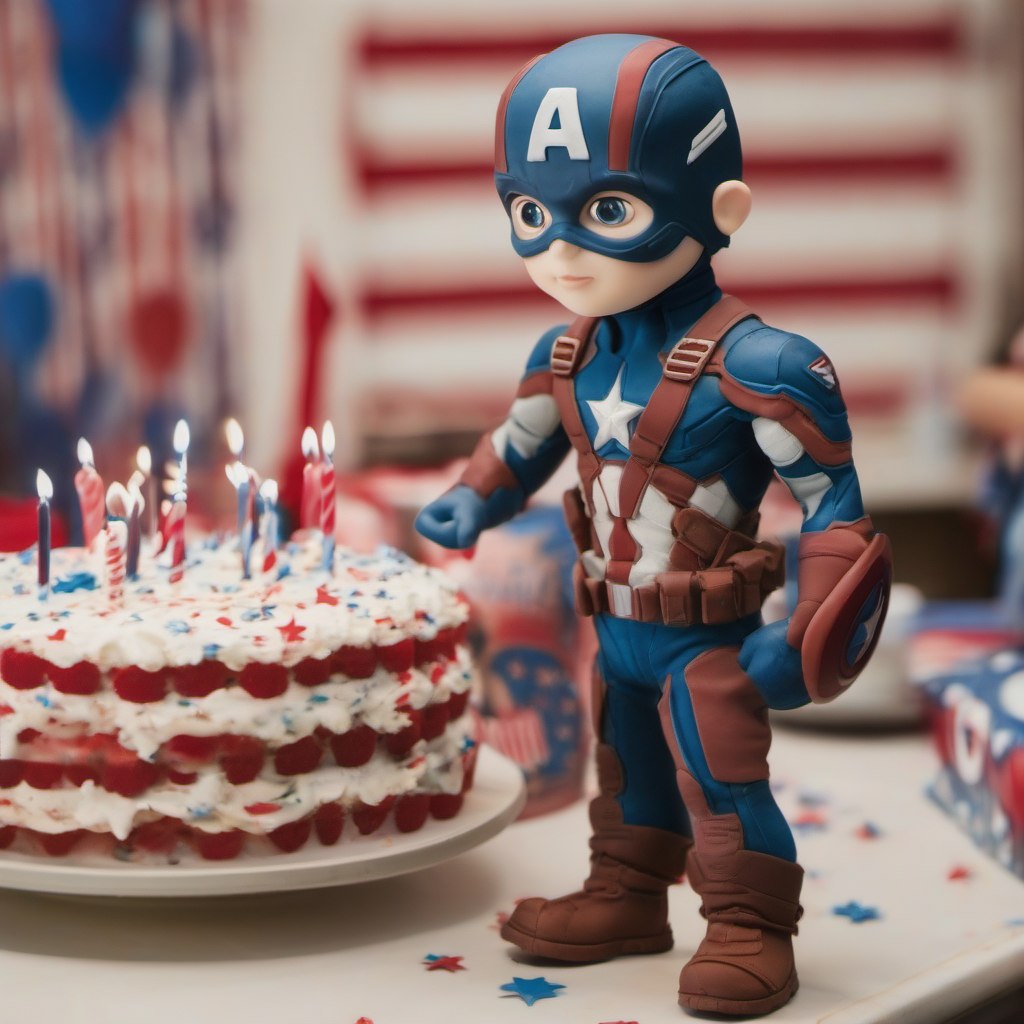}
    \caption{
    22 steps (fixed)}
    \label{fig:success_case_base}
\end{subfigure}
 \begin{subfigure}[t]{0.49\textwidth}
    \includegraphics[width=0.92\linewidth]{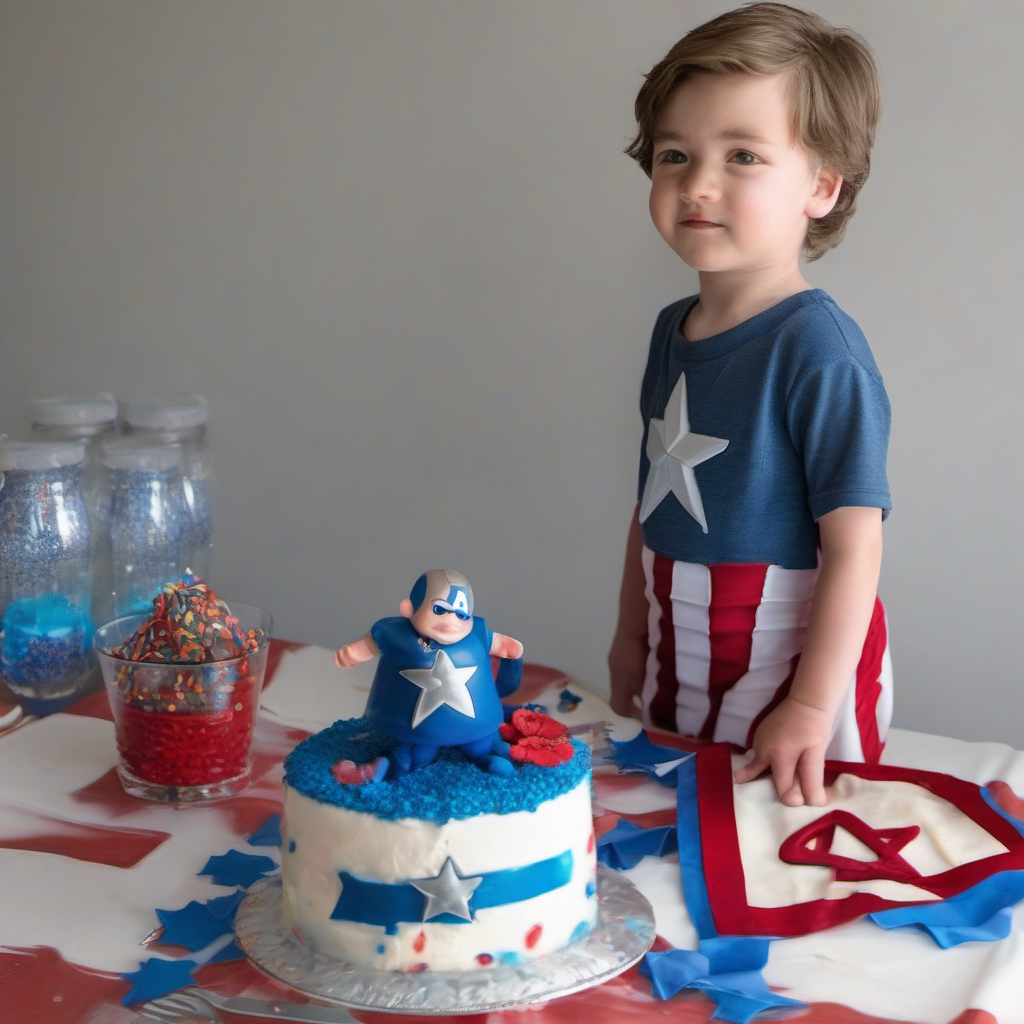}
    \caption{
    27 steps (routed)}
    \label{fig:success_case_ours}
\end{subfigure}
\end{minipage}
\hspace{0mm}
\vrule
\hspace{1mm}
\begin{minipage}[b]{.48\textwidth}
\textbf{Failure case}. 
There are two traffic signals on a metal pole, each with three light signals on them. \\
 \begin{subfigure}[t]{0.49\textwidth}
    \includegraphics[width=0.92\linewidth]{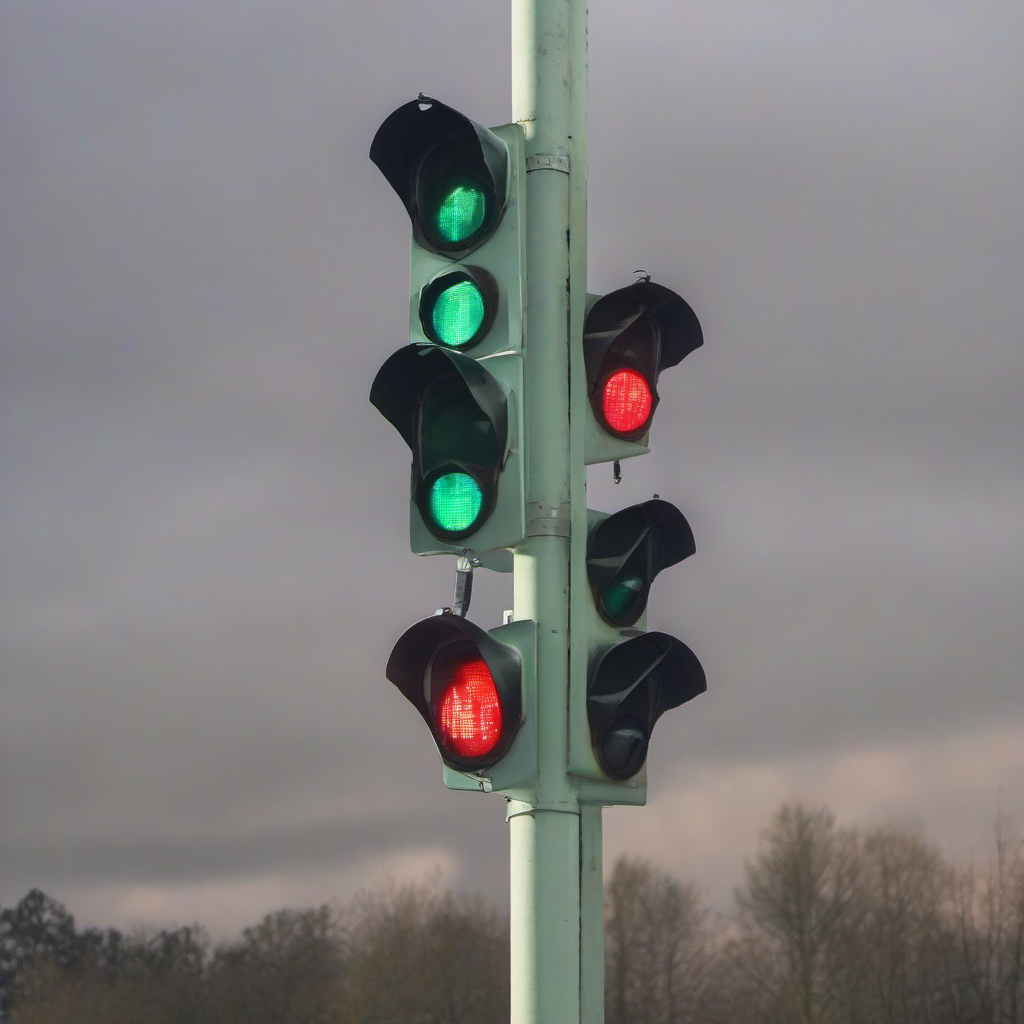}
    \caption{
    22 steps (fixed)}
    \label{fig:failure_case_base}
\end{subfigure}
\begin{subfigure}[t]{0.49\textwidth}
    \includegraphics[width=0.92\linewidth]{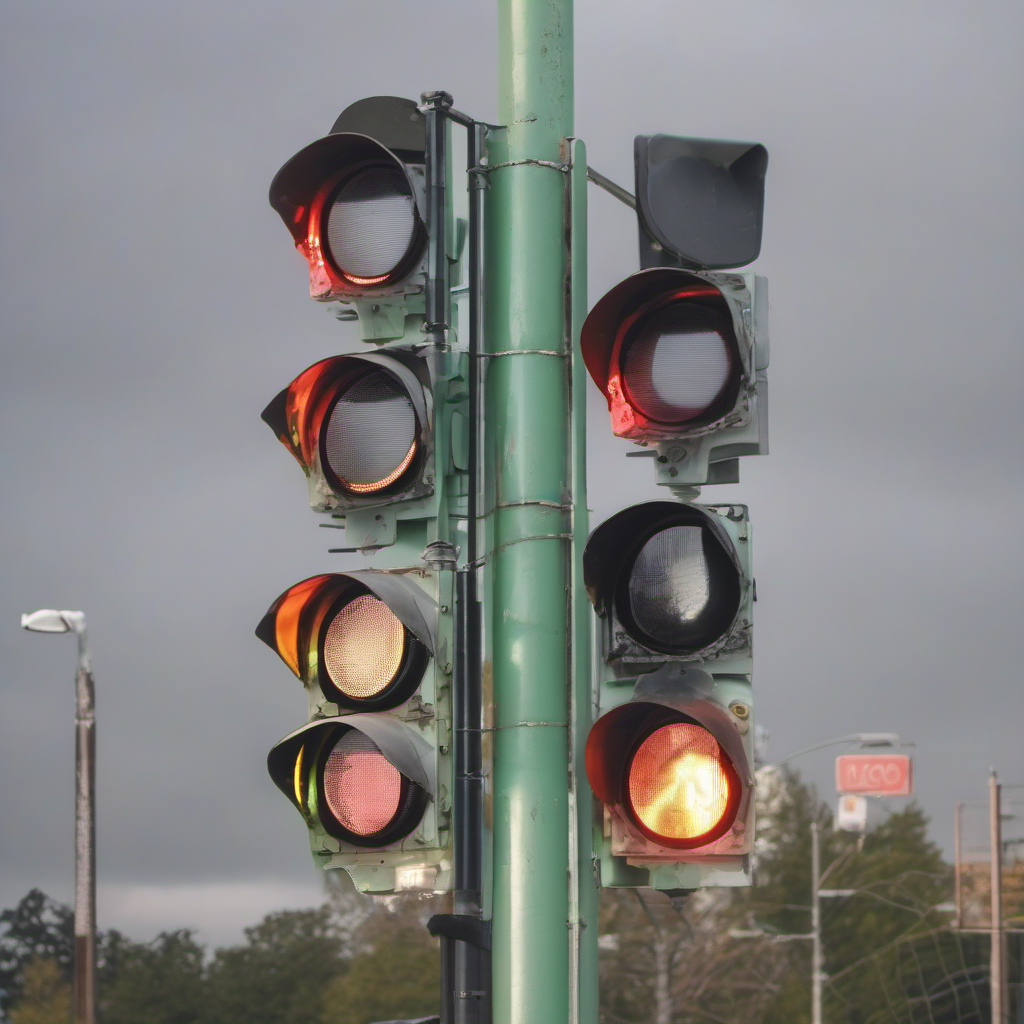}
    \caption{
    27 steps (routed)}
    \label{fig:failure_case_ours}
\end{subfigure}
\end{minipage}
\caption{Success and failure cases of the baseline SDXL with static 22 denoising steps, and our approach \textit{Proposed Transformer (SDXL+) in \Cref{fig:deferral_curve_clip}} operating at the same average cost as the baseline.  
\textbf{(a)}, \textbf{(b)}: Our approach is able to recognize the need for a larger number of denoising steps to generate an image that matches the prompt. 
\textbf{(c)}, \textbf{(d)}: Prompts that specify an exact number of objects are difficult for diffusion models in general. The number of objects may fluctuate during the denoising process, making it difficult to predict the right number of steps. 
}
\label{fig:coco_success_failure}
\vspace{-2mm}
\end{figure}

In \Cref{fig:coco_success_failure}, we analyze scenarios showing both successes and failures of our adaptive routing method (\textit{Proposed Transformer (SDXL+)} on CLIPScore metric). Specifically, we focus on cases where our method uses the same overall computational cost as the baseline (SDXL with a fixed 22 denoising steps). Within these scenarios, we consider cases where our method allocates more than 22 denoising steps, indicating that the prompts are particularly complex and require additional refinement.

For the prompt \textit{A young kid stands before a birthday
cake decorated with Captain America}, our method correctly recommends more denoising steps, as fewer would not generate accurate images. In contrast, the prompt \textit{There are two traffic signals on a metal pole, each with three light signals on them} includes an exact number of objects, a concept which both diffusion models and CLIP often struggle with 
\citep{binyamin2024count,paiss2023countclip}. Our approach accounts for this difficulty by recommending more steps than average. However, 
in this case, more denoising steps actually degrade image quality which is uncommon and ends up hurting the router performance.

We also perform a user study to compare the subset of these routing decisions with the fixed cost baseline (see Appendix \Cref{sec:supp:User_Study}). All participants rate \Cref{fig:success_case_ours} (ours) as the better image, while 14 of 19 participants select \Cref{fig:failure_case_base} (baseline) as the better image. 
\section{Cross Dataset Generalization}
To evaluate cross-dataset generalization, we conducted an experiment where our quality estimator (KNN) was trained on DiffusionDB prompts and evaluated on the prompts from COCO captioning dataset. This is a challenging test due to the significant stylistic differences between the datasets. In ~\Cref{tab:cross_data}, we report the average quality (Sharpness score) achieved by our routing approach when operating at the cost (TFLOPs) of each model in the pool. Our framework still achieves a good quality-cost trade-off even without being trained on this new prompt distribution. This suggests the estimator learns a robust understanding of prompt complexity beyond simple keyword correlation.
\begin{center}
\centering
    \label{tab:cross_data}
 \begin{tabular}{l*{5}{r}}
        \toprule
        \multicolumn{6}{l}{\textbf{Sharpness} {\color{teal}$\uparrow$}}\\
    \midrule
          &  Turbo & Lighting & DeepCache & DDIM & SDXL-100\\
        \midrule
        Ours & 0.1152 & 0.1150 & 0.1134 & 0.1100 & 0.1048\\
        Fixed & 0.1152 & 0.1069 & 0.0979 & 0.1049 & 0.1048\\
        \bottomrule
    \end{tabular}
\vspace{-2mm}
\end{center}
\section{Model Architecture}
\label{sec:supp:model_architect}
In this section, we provide more details on the architecture of the models we used for the quality estimator $\hat{\gamma}^{m}$ in Eq. $\eqref{eq:emp_rule}$. 

\paragraph{$K$-NN}

The \textsc{$K$-NN} approach provides a non-parametric way to estimate quality by retrieving and averaging the quality scores of $K$ nearest training prompts, in an appropriate embedding space. 
This method is simple, and can generalize well with sufficient data. 
As no iterative training is required for $K$-NN (besides producing a search index), it can be a suitable model when the underlying pool $\mathscr{H}$ of base models changes frequently.
We use the CLIP text encoder \cite{radford2021learning} to produce prompt embeddings 
The text encoder is used directly without fine-tuning.

\paragraph{Transformer}

Our Transformer-based estimator is built on top of the text embeddings produced by the CLIP text encoder. Specifically, for each prompt, CLIP considers the first 77 tokens and produces 77 per-token embeddings, each of 768 dimensions.
To construct our quality estimation model,  we first add two self-attention layers with position embeddings, resulting in an output sequence of 77 per-token embeddings, each of 512 dimensions.
We then add a small output head with a 2-layer linear MLP with a Sigmoid activation function on top of each of these token embeddings. 
Averaging across the tokens produces $M$ scores $\hat{\gamma}^{(1)}(\bx), \ldots, \hat{\gamma}(\bx)^{(M)}$ (see Eq. \eqref{eq:emp_rule}), each estimating the expected quality of the $m$-th model on prompt $\bx$.

We train a separate model for each of the quality metrics considered. 
In each case, the quality scores are linearly normalized across all training examples to be in [0, 1]. 
These normalized metrics are treated as ground-truth probabilities, and the model is trained by minimizing the cross-entropy losses. Only the attention layers and the MLP are trained with the frozen CLIP text encoder.

Note that all the base models except Infinity consume the CLIP text embedding as input. Thus, the cost of invoking our router is only from the extra layers added in the case of Transformer, or neighbor lookup in the case of $K$-NN. The overhead in terms of FLOPs is negligible compared to invoking SDXL. Since the Infinity baseline uses Flan-T5 instead of CLIP, this incurs an additional $\sim 13.087$ GFLOPs if Infinity is selected. To put it in perspective, calling SDXL for one prompt with 17 denoising steps would incur roughly 200 TFLOPs.

\section{Computational Resources}
\label{sec:supp:Compute_Resources}

To generate our training set (quality score per prompt), we used 50 A100/H100 80G GPU with approximately $\sim$4 days per model to generate 391,364 images for the filtered DiffusionDB prompt set (97,841 prompts) and 
less than 1 day per model for the filtered COCO prompt set (18,384 prompts). For each quality metric, we trained our Transformer with one A100 40G GPU in $\sim$2 hours for the COCO prompt set and $\sim$7 hours for the DiffusionDB prompt set.
The trained transformer (33.88M parameters, 15.61 GFLOPs) represents just 1.7\% of the size and 1.04\% of the computational cost relative to the smallest candidate, Infinity (2B parameters, 1.5 TFLOPs).
For our kNN-based router, the overhead is even more marginal, as it only requires a single-pass search over a small set of $10^3$--$10^4$ reference samples.
The trained Transformer model only takes $\sim$0.05 seconds to predict the scores for a single prompt in one A100 40G GPU. We trained KNN on the CPU in less than 1 minute with negligible inference time($<$0.01 second per prompt).


\section{Quality-Neutral Costs}
\label{sec:supp:Quality_Costs}

We provide quantitative metrics to complement the deferral curves shown in \Cref{fig:deferral_curve_clip} (COCO dataset, evaluated with CLIPScore). 
We consider the quality-neutral cost (QNC) \cite{OngAlmWu2025,JitNarRaw2025} defined as the fraction of cost required to reach the performance of a reference model.
The lower the QNC, the better because this means that our method can achieve the same performance as a reference model using a lower cost.
The QNCs of the two proposed methods are shown in the following table, where the reference is set to each of the individual model in the pool (described in \Cref{sec:exp:setting}).

\begin{center}
\resizebox{\linewidth}{!}{
\begin{tabular}{lllllll}
\toprule
Method $\quad \backslash \quad$ QNC (\%) & \textsc{Infinity} & \textsc{DDIM} & \textsc{DeepCache} & \textsc{Lighting} & \textsc{Turbo} & \textsc{SDXL100} \\
\midrule
\textit{Transformer (SDXL+)} & 100 & 2.5 & 11.1 & 6.4 & 99.7 & 2.1 \\
\textit{$K$-NN (SDXL+)} & 100 & 2.5 & 12.3 & 6.4 & 100.2 & 2.4 \\
\bottomrule
\end{tabular}
}
\end{center}

For example, a QNC of 100\% to \textsc{Infinity} indicates that our approaches would need the full cost of \textsc{Infinity} to reach its average performance. In other words, visually, the deferral curves of the two proposed methods would pass through the quality-cost operating point of \textsc{Infinity}.
As another example, the QNC of the \textit{Proposed Transformer (SDXL+)} relative to the baseline \textsc{DeepCache} is 11.1\%, indicating that \textit{Proposed Transformer (SDXL+)}  only needs 11.1\% of the cost of \textsc{DeepCache} to have the same performance.

Overall, our proposed approaches are able to match the quality of all the baselines with either a significantly lower cost, or almost the same cost.


\newpage
\section{Model Selection Rates} 
\label{sec:supp:selection_rates}

\begin{figure}[h]
\centering
\includegraphics[width=0.53\linewidth]{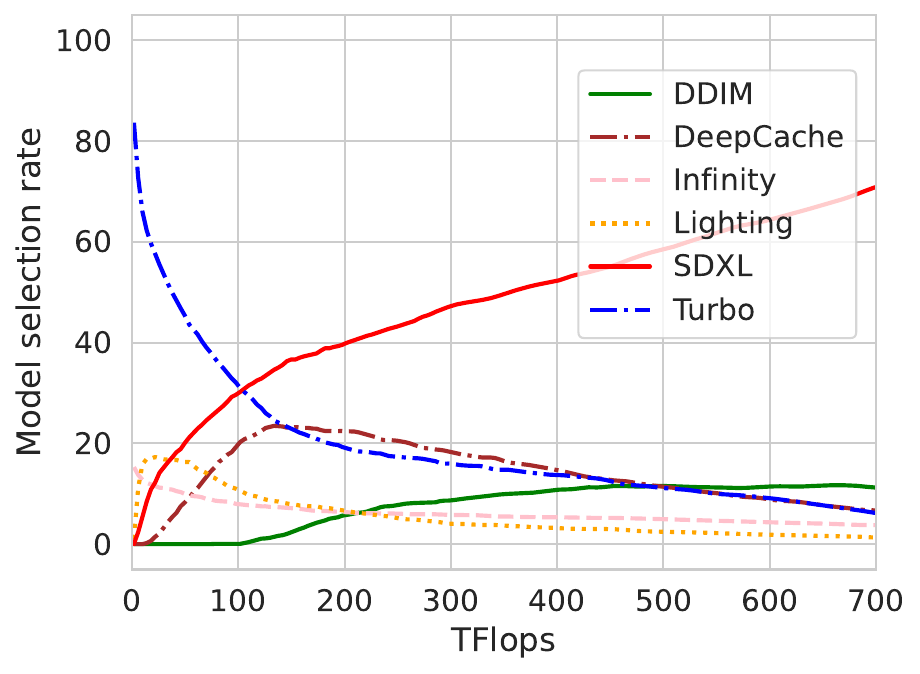}
\caption{ The rate at which each choice in the candidate routing pool is selected by \textit{Proposed Transformer (SDXL+)}  in \Cref{fig:deferral_curve_clip}.
Our approach is able to adaptively mix and match different model choices throughout the cost range.
}
\label{fig:selection_rates_CLIP}
\vspace{-3mm}
\end{figure}

\Cref{fig:selection_rates_CLIP} shows the rate at which each choice in $\mathscr{H}$ is selected by \textit{Proposed Transformer (SDXL+)}  in \Cref{fig:deferral_curve_clip}.
All the 12 candidate diffusion steps offered by the base SDXL model are collapsed into one curve for clarity.
We observe that,  when the cost budget is large, the router increasingly allocates resources to the full SDXL model. 
On the other hand, in the lower cost range, \textsc{Turbo} is the prominent choice, as it provides a good balance between cost and quality.
This analysis shows that our router is able to adaptively mix and match different choices throughout the cost range to achieve a good quality-cost trade-off.

\section{$K$-NN Parameter Selection}
\label{sec:supp:KNN_parameter}

\begin{figure}[H]
\centering
\includegraphics[width=0.48\linewidth]{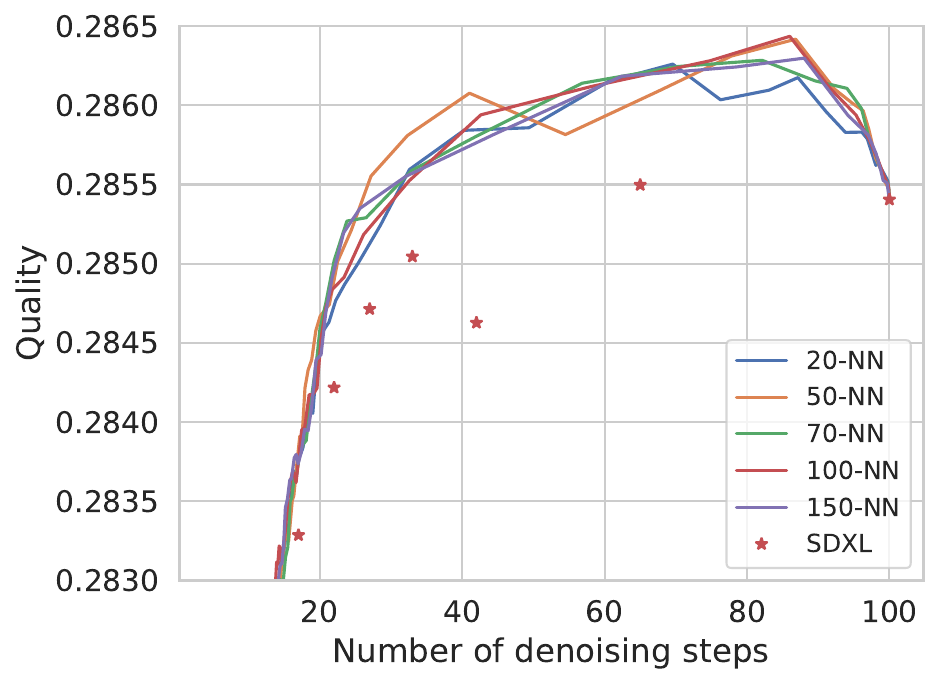}
\caption{ Performance of the proposed $K$-NN-based routing model on a validation subset drawn from COCO.
}
\label{fig:ablation_knn}
\end{figure}

\Cref{fig:ablation_knn} shows the deferral curve when using $K$-NN as a quality estimator at various values of $K$. As shown in the figure, the routing performance (on a validation set of 828 prompts drawn from the COCO dataset) is similar across a wide range of $K$ values. We set $K = 100$ for our final model.

\section{User Study}
\label{sec:supp:User_Study}

Here we provide details on the user study to evaluate our routing decision.
For this qualitative analysis, we consider the same trained router shown in \Cref{fig:deferral_curve_clip} as \textit{Proposed Transformer (SDXL+)}.
Our reference baseline is the SDXL model with the number of denoising steps set to 22; this setting results in a per-prompt cost of 263.3 TFLOPs.
For a fair comparison,  we accordingly consider the operating point of our method that has the same average cost as this baseline by adjusting $\lambda$ in Eq. \eqref{eq:emp_rule}.
Of all the prompts in the test set used in \Cref{fig:deferral_curve_clip}, we consider a random subset of 100 prompts where our method does not select SDXL with 22 denoising steps as the routed decision; this filtering is done to facilitate a contrast between the two approaches.
We proceeded to recruit 19 participants through the \textit{Prolific} platform for a human preference analysis (\url{https://www.prolific.com}).
We run a two-alternative forced-choice (2AFC) study to measure participants' preference for images produced by both approaches.
During each trial, each participant is presented with a text prompt, and two test images produced by the two approaches.
The participant is instructed to select the image that better matches the prompt.

\textbf{Protocol}
For each trial, a prompt is shown first to each participant. 
Two randomized-order images are then presented on participants' screens: 1) image from our \textit{Proposed Transformer (SDXL+)} (in \Cref{fig:deferral_curve_clip}), and 2) image produced by the baseline SDXL at fixed 22 denoising steps.
Note that the produced image may be from a non-SDXL model (e.g., Turbo) since our approach may route to other baseline models described in \Cref{sec:exp:setting}.

\begin{figure*}[h!]
\centering
\includegraphics[width=0.6\linewidth]{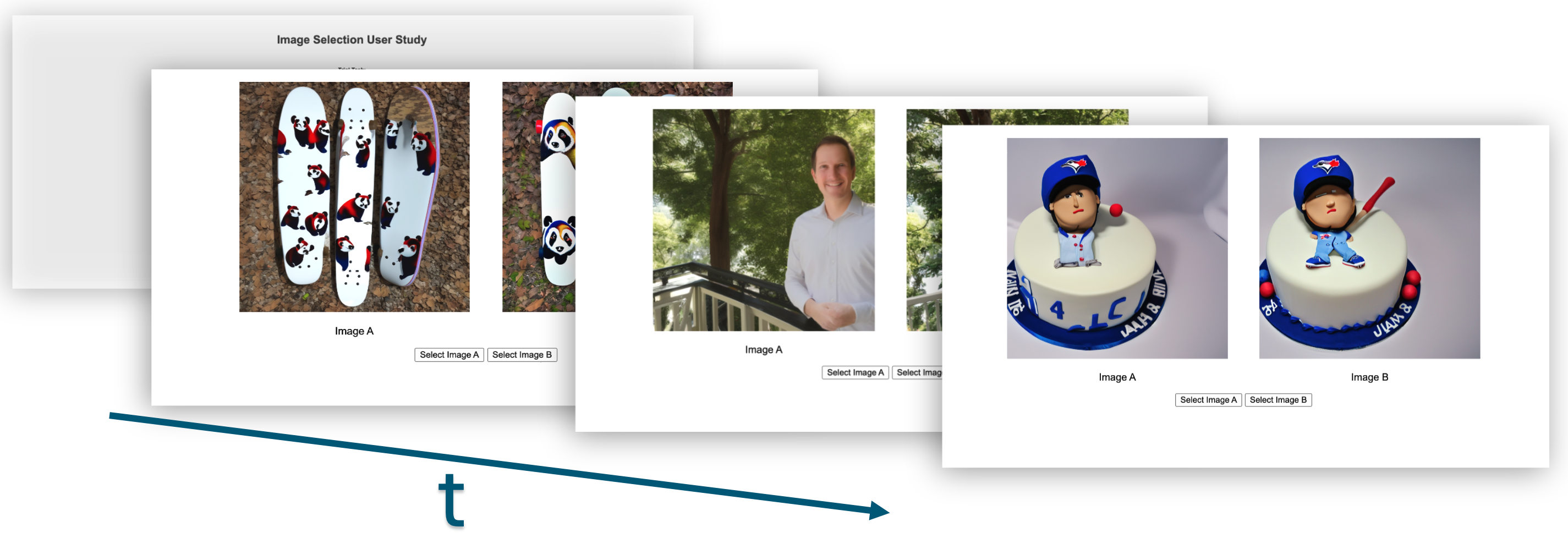}
\Caption{User study protocol Ours vs. SDXL with static 22 denoising steps.}
{
In each user study trial, the participant will see two images: Image A and Image B. The task is to select the image that match better with the text prompt provided. The participant needs to click on the button below or press the keyboard to choose A/B.
}
\label{fig:supp:protocol2}
\Description{}
\vspace{-4mm}
\end{figure*}

Participants were instructed to select the image that matches the input text prompt better.
Each participant will be assigned 100 trials (100 prompts) in total, with prompts randomly sampled from the COCO dataset described in \cref{sec:method:data}. Our crowdsourcing user study protocol is visualized in \Cref{fig:supp:protocol2} as a sequence of web pages that will be shown to participants with example stimuli. 

\textbf{Additional Results}
\begin{figure}[h!]
\centering
\includegraphics[width=0.62\linewidth]{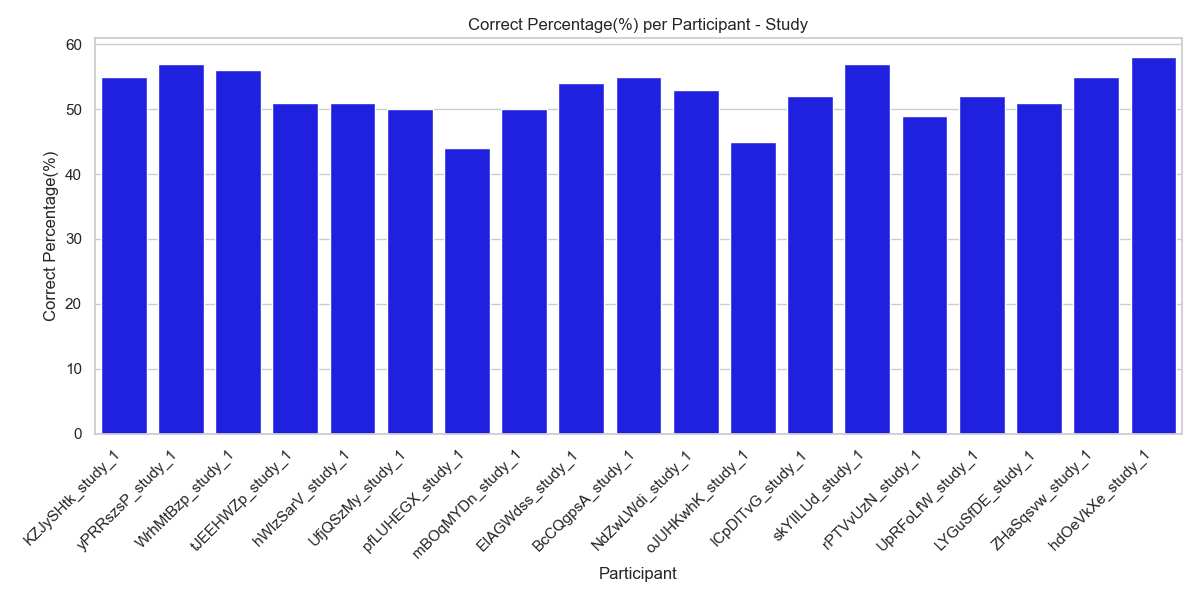}
\caption{The rate at which each participant prefers our suggested image over the image produced by the baseline SDXL with 22 denoising steps.
}
\label{fig:user_study}
\vspace{-5mm}
\end{figure}

\Cref{fig:user_study} shows the rate, in percentage, at which each participant selects our \textit{Proposed Transformer (SDXL+)}. Here, we define \textbf{percentage selection} as the proportion of trials in which ours is selected. 
We note that for Stable Diffusion XL, many images generated from 22 denoising steps already show reasonable results with hard-to-notice artifacts, which can limit the perceivable differences. 
In the main paper, we highlight that the majority of participants only agreed on those most notable cases. On average, the participants prefer ours 52.37\% of the time.
\section{Comparison with DeepCache on Quality-Cost Trade-off}
\label{sec:supp:deepcache_comparison}
\begin{figure}[hbt!]
\centering
\includegraphics[width=0.53\linewidth]{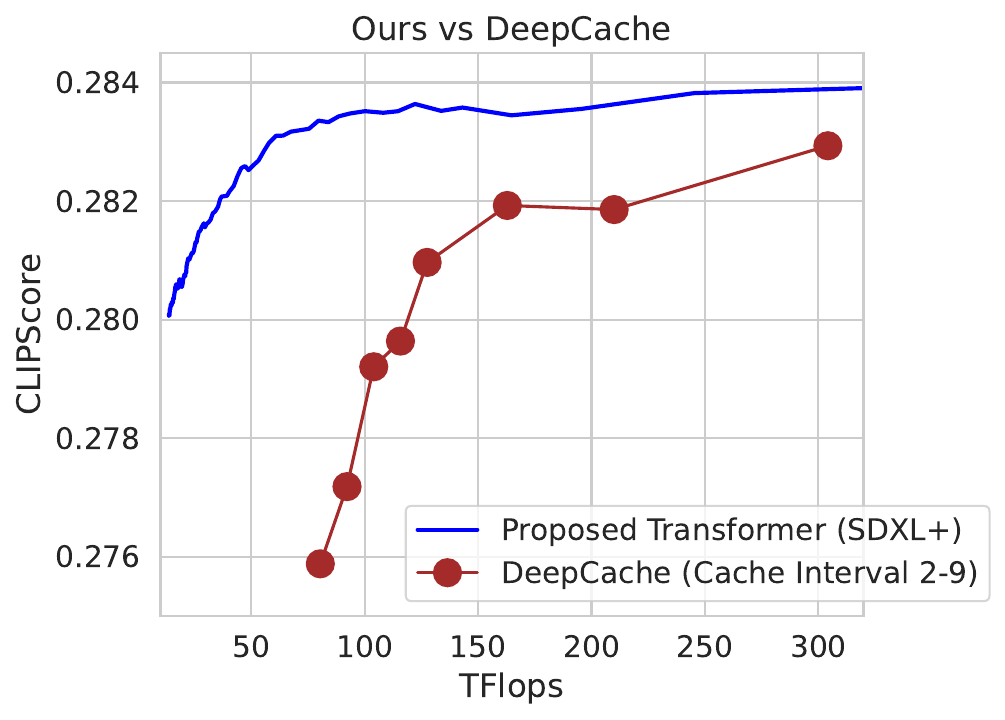}
\caption{ CLIPScore-TFlops trade off comparison with Deepcache at various cache interval on the test set of COCO dataset
}
\label{fig:deepcache_comparison}
\vspace{-3mm}
\end{figure}

\Cref{fig:deepcache_comparison} compares our adaptive routing method (\textit{Proposed Transformer (SDXL+)}) on CLIPScore with the DeepCache approach \cite{ma2023deepcache} (on SDXL model at 50 denoising steps). DeepCache caches intermediate activations at predefined intervals (cache intervals) to balance image quality and computational cost. Varying the cache interval enables different quality–cost trade-offs, which can then be compared with our method on a deferral curve. As shown in \Cref{fig:deepcache_comparison}, our method, which adaptively utilizes multiple models, consistently surpasses fixed-interval caching methods across all computational costs. Note that our adaptive routing strategy can also incorporate any DeepCache configurations to even further enhance performance.

\section{Statistical Significance of COCO Results}


Recall that \Cref{fig:main_plot} shows the deferral curves of our methods on COCO dataset.
To give a more precise view on the performance improvement, in \Cref{tab:coco_quan}, we report average quality scores attained by our \emph{Proposed K-NN (SDXL+)} at the same costs as the individual models in the pool.
We observe that at the operating cost of each individual model in the pool, our approach is able to deliver a higher average quality score (as measured by CLIPScore and Sharpness). In most cases, the gains are statistically significantly better as warranted by the Welch's t-test.
Note that at the two extreme ends of operating costs (i.e., calling the cheapest and most expensive models, respectively), any routing approach necessarily reduces to trivial routing:  sending all prompts to one model.
It follows that, at an extreme operating point (either at the lowest possible or highest possible cost), the average quality achieved must be exactly the same as that of the individual model at that end point.

\begin{table}[H]
  \centering
  \caption{Quality–cost breakdown for \emph{Proposed K-NN (SDXL+)} presented in  \Cref{tab:coco_quan} (on COCO dataset). An entry in bold text indicates that, with the same cost, our approach is statistically significantly better than the corresponding individual model (Welch's t-test at significance level $\alpha = 0.05$).}
  \label{tab:coco_quan}
  \resizebox{\linewidth}{!}{
  \begin{tabular}{l*{9}{r}}
    \toprule
    \midrule

    \multicolumn{10}{l}{\textbf{CLIPScore} \cite{radford2021learning} {\color{teal}$\uparrow$}}\\
    \midrule
    Ours  
      & 0.2672 & 0.2801 & \textbf{0.2817} & \textbf{0.2830} & \textbf{0.2832} & \textbf{0.2832} &  \textbf{0.2832} & 0.2830 & 0.2822 \\
    Fixed 
      & 0.2672 & 0.2798 & 0.2749 & 0.2773 & 0.2820 & 0.2805 & 0.2810      & 0.2816 & 0.2822 \\

    \midrule
    \multicolumn{10}{l}{\textbf{Sharpness} {\color{teal}$\uparrow$}}\\
    \midrule
    Ours  
      & 0.1061 & 0.1161 & \textbf{0.1166} & \textbf{0.1163} & \textbf{0.1160} & \textbf{0.1158} & \textbf{0.1132} & \textbf{0.1121} & 0.1048 \\
    Fixed 
      & 0.1061 & 0.1152 & 0.1070        & 0.0997 & 0.0979 & 0.0858 & 0.0995 & 0.1049 & 0.1048 \\
   
    \midrule
    \midrule
    Model 
      & Infinity & Turbo    & Lighting & SDXL-9   & DeepCache & SDXL-22  & SDXL-42  & DDIM     & SDXL100   \\
      Cost 
      &1.50 &	1.54 &	23.92 &	107.73 &	210.00 &	263.34 &	502.74 &	598.50 &	1197.00   \\
    \bottomrule
  \end{tabular}
  }
\end{table}

\section{Additional Deferral Curve} 
\label{sec:supp:diffdb_deferral_curve}
\begin{figure}[htbp]
  \centering
  
  \begin{subfigure}{.4\linewidth}
    \includegraphics[width=\linewidth]{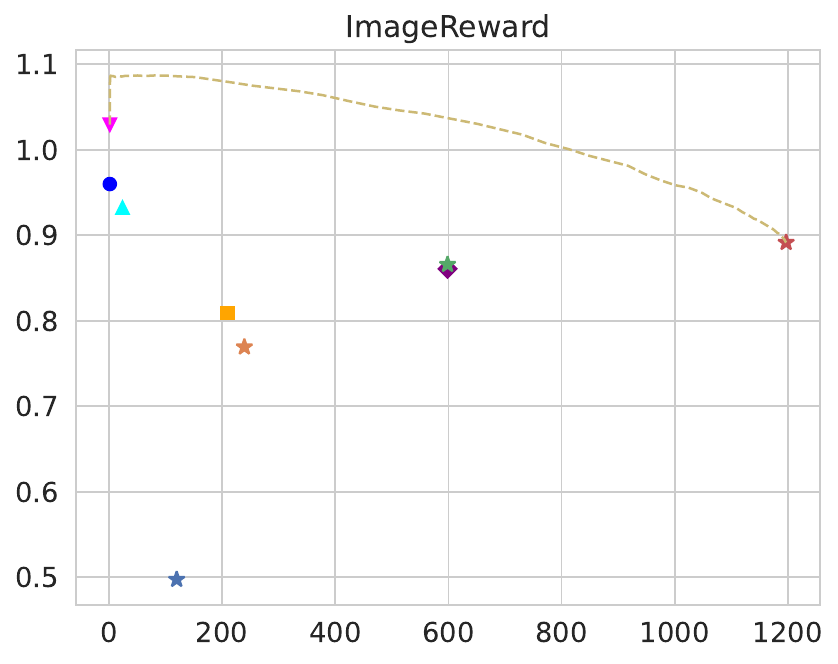}
  \end{subfigure}
  \begin{subfigure}{.4\linewidth}
    \includegraphics[width=\linewidth]{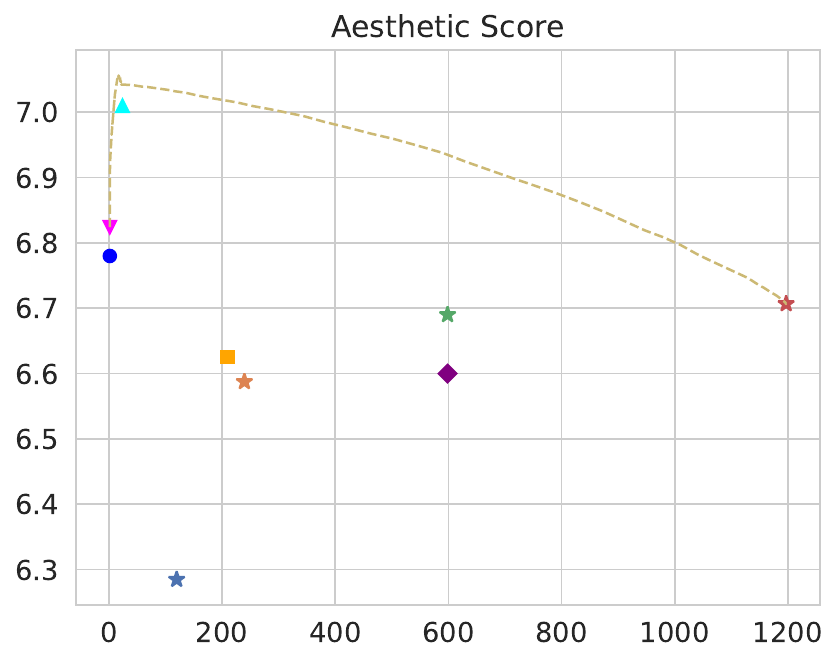}
  \end{subfigure}
  \vspace{4pt}            
  
  \begin{subfigure}{.4\linewidth}
    \includegraphics[width=\linewidth]{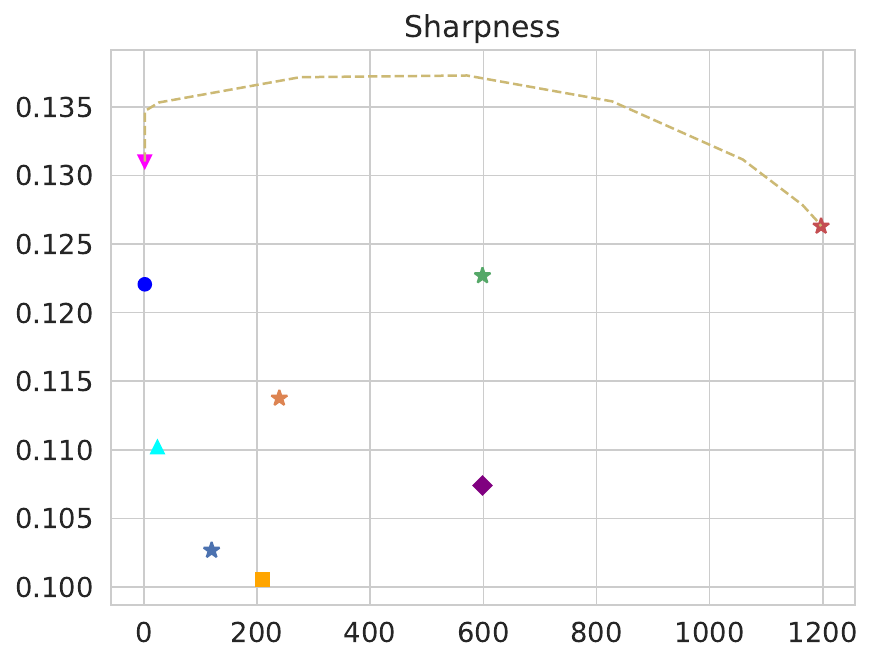}
  \end{subfigure}
  \begin{subfigure}{.4\linewidth}
    \includegraphics[width=\linewidth]{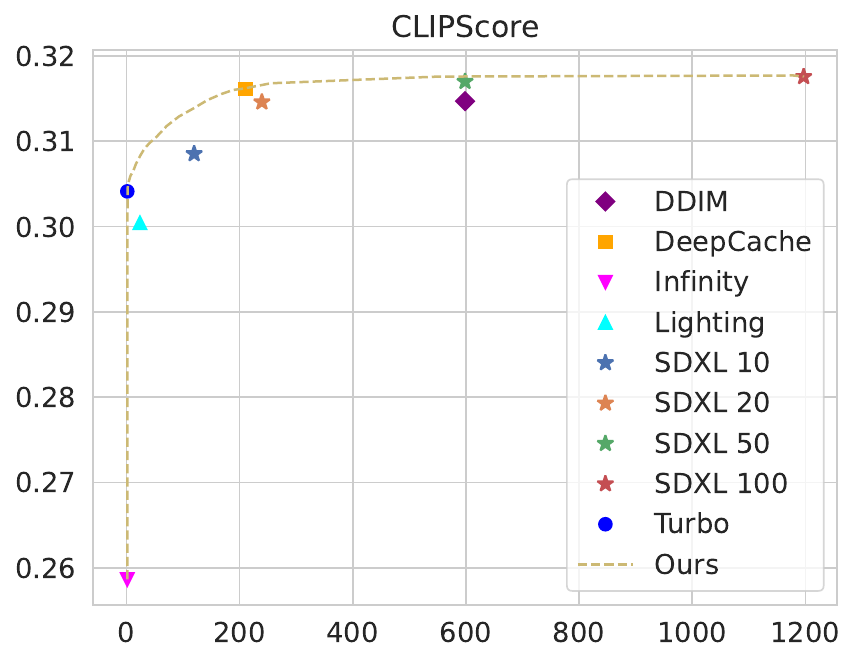}
  \end{subfigure}
  \caption{Deferral Curves of our \textit{Proposed Transformer (SDXL+)} on DiffusionDB dataset. Our approach exceeds the quality of fix-step text-to-image models in all quality metrics (ImageReward, Aesthetic, Sharpness, and CLIPScore).}
  \label{fig:deferral_diffdb}
\end{figure}

Here we show the complete deferral curves for the four quality metrics--CLIPScore, Sharpness, ImageReward, and Aesthetic Score, on the test set of DiffusionDB dataset in \Cref{fig:deferral_diffdb}. These curves complement the fixed-cost comparison in \Cref{tab:quantiative_results} by showing the changes in quality score across the entire cost spectrum. Here we can clearly see how our adaptive routing consistently achieves a higher quality score than the fixed-model baselines at every computational budget. \\

\clearpage
\section{Estimation Errors and Quality-Cost Trade-offs}
\label{sec:noisy_oracle}

\begin{figure}[h]
    \centering
    \includegraphics[width=0.75\linewidth]{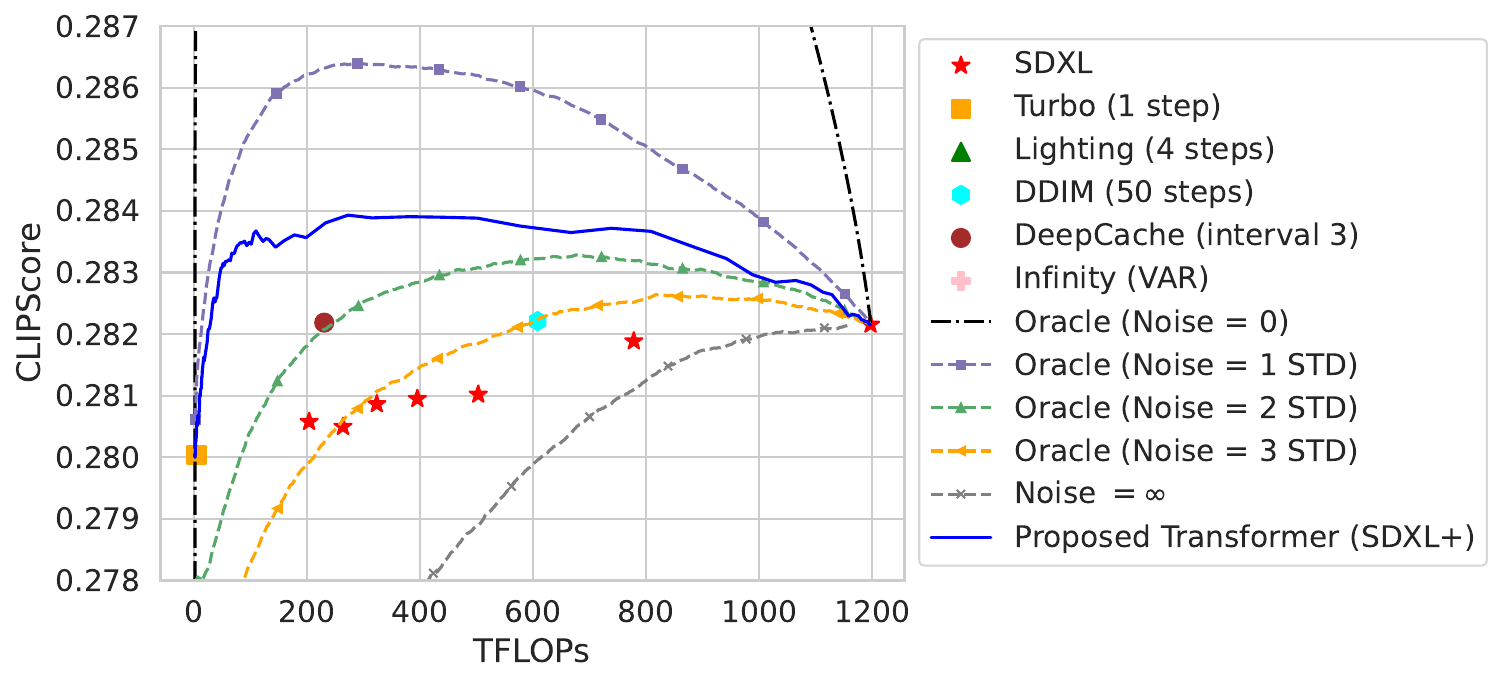}
    \caption{The deferral curves as presented in \Cref{fig:deferral_curve_clip}, supplemented with baselines derived from the oracle routing rule with noise added to simulate routing errors. See \Cref{sec:noisy_oracle} for details.}
    \label{fig:noisy_oracle_curves}
\end{figure}

\begin{table}[ht]
    \centering
    \resizebox{\textwidth}{!}{%
    \begin{tabular}{lcccccccc}
    \toprule
     \multicolumn{6}{l}{\textbf{CLIPScore} {\color{teal}$\uparrow$}}\\
     \midrule
                   & Infinity & Turbo  & Lighting & SDXL-9  & DeepCache & SDXL-22 & DDIM   & SDXL-65 \\ 
    \midrule
  Ours & .2672 & .2800 & .2820 & .2836 & .2837 & .2840 & .2838 & .2837 \\
    Fixed & .2672 & .2798 & .2749 & .2773 & .2820 & .2805 & .2816 & .2819 \\
    \midrule \midrule
 Oracle & .2672 & .2855 & .2928 & .2973 & .2984 & .2986 & .2969 & .2944 \\
Oracle (Noise = 1 STD) &.2672 & .2806 & .2829 & .2854 & .2862 & .2864 & .2860 & .2851 \\
Oracle (Noise = 2 STD) & .2672 & .2780 & .2782 & .2806 & .2819 & .2823 & .2832 & .2832 \\
Oracle (Noise = 3 STD) &.2672 & .2768 & .2763 & .2784 & .2800 & .2806 & .2822 & .2825 \\
Random (Noise = $\infty$) & .2672 & .2727 & .2711 & .2733 & .2756 & .2763 & .2799 & .2811 \\
    \midrule
    \rowcolor{gray!15}
    TFLOPs         & 1.5      & 1.54   & 23.92    & 107.73  & 210       & 263.34  & 598.5  & 778.05  \\ 
    \bottomrule
    \end{tabular}%
    }
    \caption{Average quality scores of the routing approaches presented in \Cref{fig:noisy_oracle_curves}. 
        }
    \label{tab:routing_noise}
\end{table}

In this section, we discuss the relationship between estimation errors of a quality metric, and the resulted quality-cost trade-off. 
Specifically, the goal is to quantify the degradation of the routing performance from the oracle routing rule, as estimation errors increase.
To demonstrate this, we consider the same experimental setup as used in \Cref{fig:deferral_curve_clip} i.e., with COCO as the dataset, and with CLIPScore as the image quality metric.

\paragraph{Plug-in estimate of the oracle rule}
We start with a plug-in empirical estimator $\hat{r}^*$ of the optimal (oracle) routing rule in \Cref{prop:opt_router}. Recall that the optimal routing rule in   \Cref{prop:opt_router} is given by 
\begin{align*}
r^{*}(\bx) & =\arg\max_{m\in[M]}\mathbb{E}\left[q(\bx,h^{(m)}(\bx))\mid\bx\right]-\lambda\cdot c^{(m)}.
\end{align*}
Recall from \Cref{sec:opt_rule} that $\hat{y}_{i,m}$ denotes the empirical estimate of  $\mathbb{E}\left[q(\bx_i,h^{(m)}(\bx))\mid\bx_i \right]$).
Define $\hat{\by}_i \defeq (\hat{y}_{i, 1}, \ldots, \hat{y}_{i, M})$.
For a labeled example $(\bx_i, \hat{y}_i)$, the optimal routing rule can thus be estimated as
\begin{align}
    \hat{r}^{*}(\bx_i) & = \arg\max_{m\in[M]} \hat{y}_{i, m} -\lambda\cdot c^{(m)}.
    \label{eq:emp_oracle_rule}
\end{align}
This data-based oracle rule is directly applicable to test examples in the test set without require any estimation. The resulting deferral curve is denoted by ``Oracle (Noise = 0)'' in \Cref{fig:noisy_oracle_curves}. Evidently, this routing rule exhibits an excellent quality-cost trade-off compared to other routing approaches. Indeed, it makes use of the ground-truth quality label $\by$ to make a routing decision. By construction, no other routing methods can give a better trade-off curve than this deferral curve (on this specific dataset).
 We emphasize that in practice it is extremely challenging to realize a quality-cost operating point that is close to this oracle routing rule. See, for instance, Figure 3 and Figure 4 in \citet{HuBieLi2024}  for the performance gap to the oracle in the context of LLM routing (i.e., not routing to text-to-image models, as considered in our work). Nevertheless, the oracle curve serves as an upper bound on the trade-off achievable by any routing methods. 

\paragraph{Noisy oracle}

We now consider adding noise to the routing rule in \eqref{eq:emp_oracle_rule} to examine how noise affects the quality-cost trade-off:
\begin{align}
    \hat{r}^{*}_\beta(\bx_i) & = \arg\max_{m\in[M]} \hat{y}_{i, m} + \beta \cdot \mathrm{STD}_m \cdot g_{i, m} -\lambda\cdot c^{(m)},
    \label{eq:emp_noisy_oracle_rule}
\end{align}
where
\begin{itemize}
    \item $\beta \ge 0$ controls the strength of Gaussian noise to add;
    \item $\mathrm{STD}_m \defeq \sqrt{\frac{1}{N_\mathrm{te}} \sum_{n=1}^{N_{\mathrm{te}}} (y_{n,m} - \bar{y}_m)^2}$, and 
    $\bar{y}_m \defeq \frac{1}{N_\mathrm{te}} \sum_{n=1}^{N_{\mathrm{te}}}  y_{n,m} $;
    \item $N_{\mathrm{te}}$ denotes the number of test examples; and
    \item $g_{i,m} \stackrel{\mathrm{i.i.d.}}{\sim} \mathcal{N}(0, 1)$ is an independent realization from the standard normal distribution.
\end{itemize}

Clearly, if $\beta=0$, we recover the empirical oracle rule in \eqref{eq:emp_oracle_rule}. The parameter $\beta$ represents the amount of Gaussian noise added to the quality scores, in the unit of the standard deviation of the quality scores of the respective model.

In \Cref{fig:noisy_oracle_curves} and \Cref{tab:routing_noise}, we present the performance of the noisy routing rule \label{eq:emp_noisy_oracle_rule} with $\beta \in \{1, 2, 3, \infty\}$. When $\beta = \infty$, we simply perform random routing, which gives a poor trade-off. 
It can be seen that our proposed the deferral curve of our proposed router lies between the curves of the oracle with $\beta=1$ and the oracle with $\beta=2$.
Roughly, this means that our router has a similar performance to an oracle router where the per-model ground-truth scores are corrupted with independent Gaussian noise at a factor $\beta \in (1, 2)$ of their standard deviations.


\end{document}